\documentclass[journal, twoside]{IEEEtran}
\usepackage{amsmath,amsfonts}
\usepackage{algorithm}
\usepackage{algpseudocode}
\usepackage{array}
\usepackage[caption=false,font=normalsize,labelfont=sf,textfont=sf]{subfig}
\usepackage{textcomp}
\usepackage{stfloats}
\usepackage{url}
\usepackage{verbatim}
\usepackage{graphicx}
\usepackage{cite}
\usepackage{hyperref}
\usepackage{booktabs}
\usepackage{xcolor}
\usepackage{eso-pic}
\AddToShipoutPictureFG*{%
  \AtPageLowerLeft{%
    \put(\LenToUnit{0.5\paperwidth},\LenToUnit{0.15in}){%
      \makebox[0pt][c]{\parbox[b]{0.85\paperwidth}{\centering\scriptsize
        \copyright~2026 IEEE. Personal use of this material is permitted. Permission from IEEE must be obtained for all other uses, in any current or future media, including reprinting/republishing this material for advertising or promotional purposes, creating new collective works, for resale or redistribution to servers or lists, or reuse of any copyrighted component of this work in other works.\\
        This is the accepted version of the article published in IEEE Transactions on Intelligent Vehicles, DOI: \href{https://doi.org/10.1109/TIV.2026.3750011}{10.1109/TIV.2026.3750011}.}}}}}
\begin{document}

\title{Learning-based Autonomous Oversteer Control and Collision Avoidance}

\author{Seokjun Lee and Seung-Hyun Kong*,~\IEEEmembership{Senior Member,~IEEE}
\thanks{*Corresponding author (email: skong@kaist.ac.kr)}
\thanks{Seokjun Lee and Seung-Hyun Kong are with the CCS Graduate School of Mobility, Korea Advanced Institute of Science and Technology, Daejeon, Korea, 34051 (email: \{seokjunlee, skong\}@kaist.ac.kr)}}

\markboth{IEEE Transactions on Intelligent Vehicles (Accepted), DOI: 10.1109/TIV.2026.3750011}{LEE \MakeLowercase{\textit{et al.}}: Learning-based Autonomous Oversteer Control and Collision Avoidance}


\maketitle

\begin{abstract}
Oversteer, wherein a vehicle’s rear tires lose traction and induce unintentional excessive yaw, poses critical safety challenges. Failing to control oversteer often leads to severe traffic accidents. Although recent autonomous driving efforts have attempted to handle oversteer through stabilizing maneuvers, the majority rely on expert-defined trajectories or assume obstacle-free environments, limiting real-world applicability. This paper introduces a novel end-to-end (E2E) autonomous driving approach that tackles oversteer control and collision avoidance simultaneously. Existing E2E techniques, including Imitation Learning (IL), Reinforcement Learning (RL), and Hybrid Learning (HL), generally require near-optimal demonstrations or extensive experience. Yet even skilled human drivers struggle to provide perfect demonstrations under oversteer, and high transition variance hinders the accumulation of sufficient data. Hence, we present Q-Compared Soft Actor-Critic (QC-SAC), a new HL algorithm that effectively learns from immature demonstration data and adapts rapidly to new conditions. To evaluate QC-SAC, we introduce a benchmark inspired by real-world driver training: a vehicle encounters sudden oversteer on a slippery surface and must avoid randomly placed obstacles ahead. Experimental results show that QC-SAC attains near-optimal driving policies, significantly surpassing state-of-the-art IL, RL, and HL baselines. Hardware-in-the-loop simulation (HILS) validation with physical steering hardware further demonstrates that QC-SAC can operate beyond a software-only simulation, providing an intermediate validation step toward real-vehicle implementation. To the best of our knowledge, this is the first work that achieves safe autonomous control of vehicles in oversteer while simultaneously avoiding obstacles.
\end{abstract}


\section{Introduction}
\begin{figure}[t!]
\centering
\subfloat[\label{fig:oversteer}]{
    \includegraphics[width=0.48\linewidth]{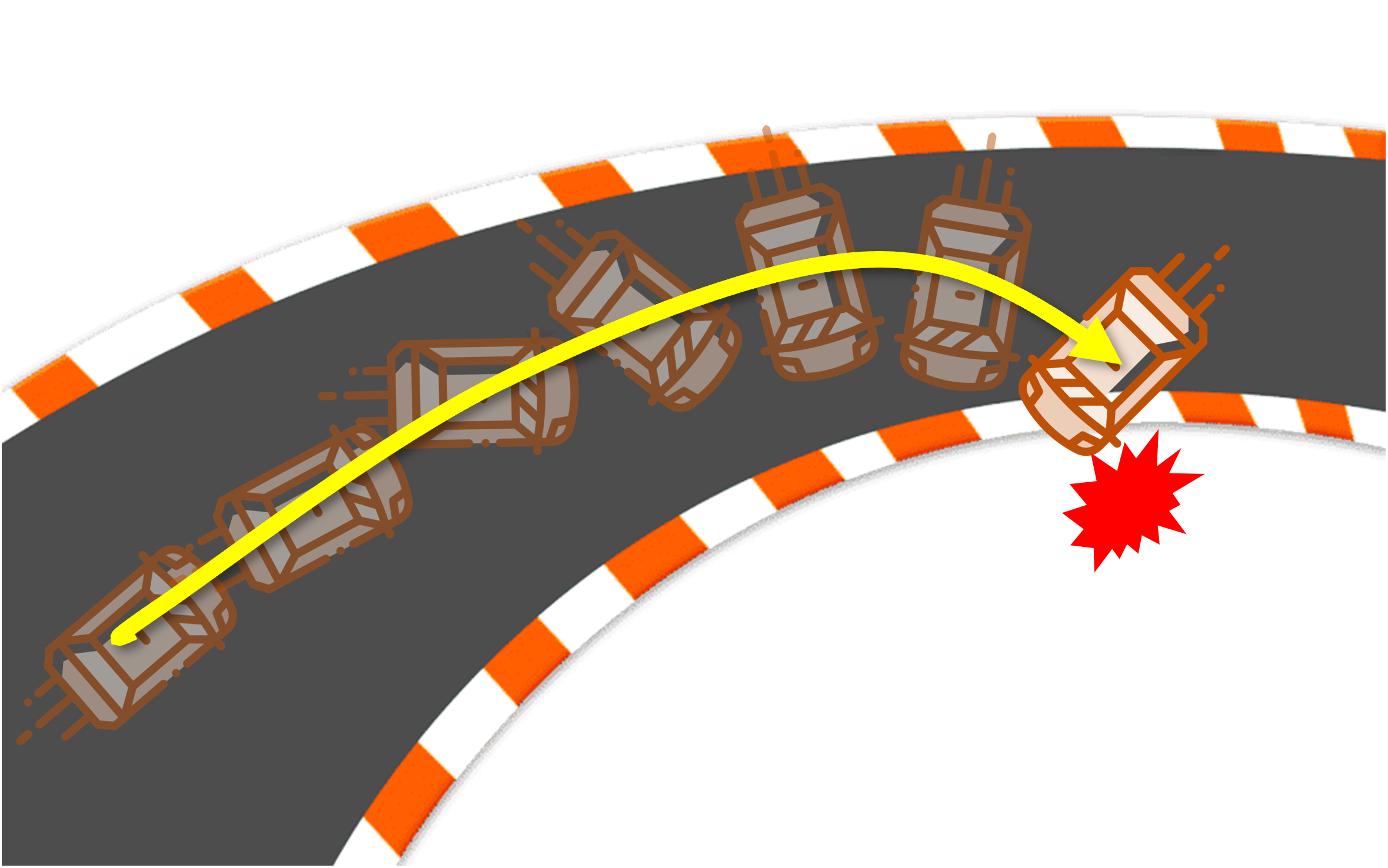}}
\subfloat[\label{fig:understeer}]{
    \includegraphics[width=0.48\linewidth]{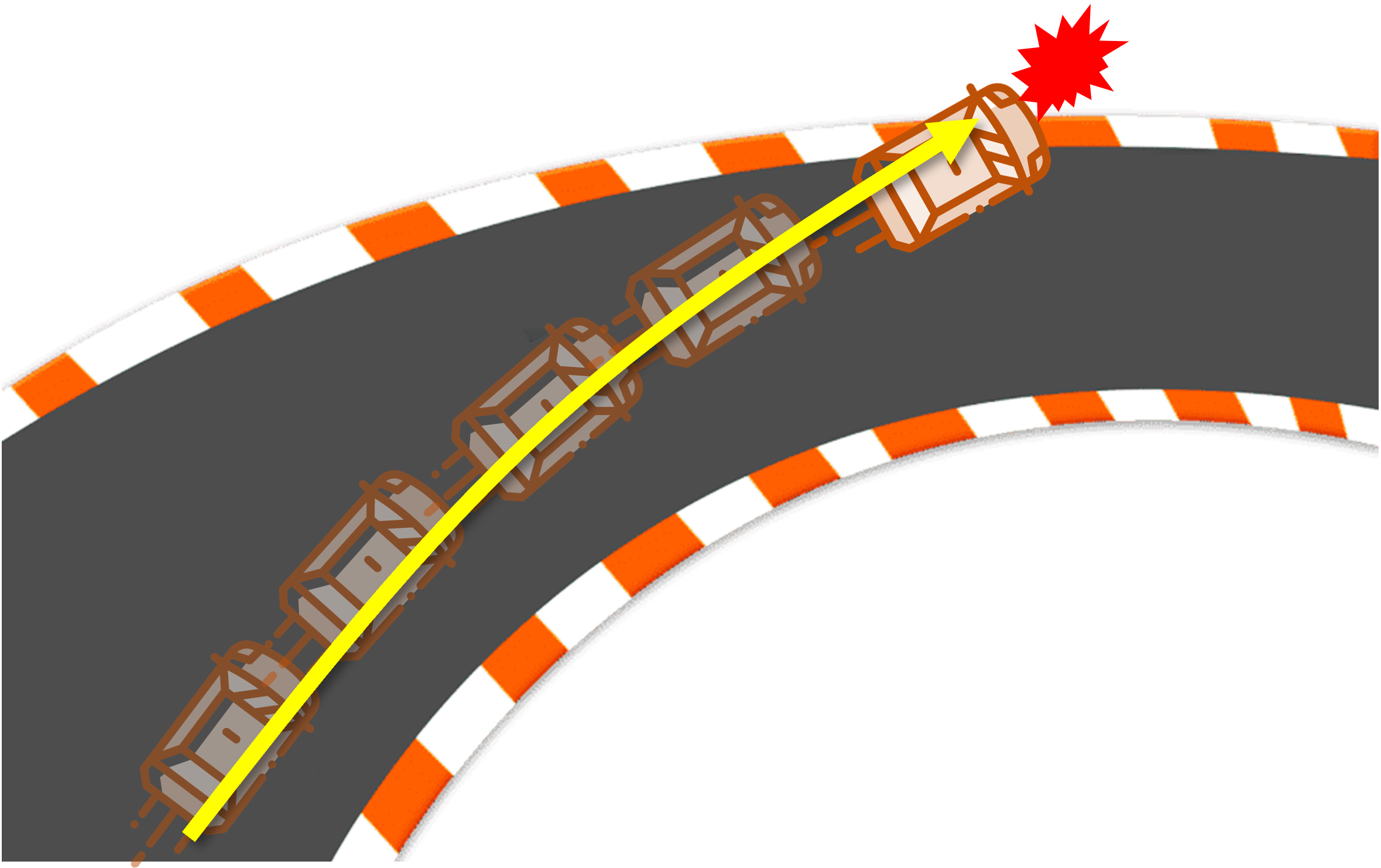}}
\caption{Vehicle oversteer and understeer. (a) Oversteer: the rear tires lose grip, and the vehicle rotates more than intended. (b) Understeer: the front tires lose grip, and the vehicle turns less than expected.}
\label{fig:oversteer+understeer} 
\end{figure}

\IEEEPARstart{W}{hen} slip occurs between a vehicle's tires and the road surface, causing the tires to skid, the vehicle becomes highly unstable. Particularly, when the rear tires lose traction, the vehicle rotates more than the driver intends and heads in an unintended direction; this phenomenon is known as oversteer (Refer to Fig.~\ref{fig:oversteer}). Oversteer primarily occurs due to causes such as reduced road friction caused by road icing or hydroplaning, tire slippage from sharp steering or pedal manipulation beyond the vehicle’s limits, or rear-end collisions. Controlling oversteer requires both appropriate pedal manipulation and adequate counter-steering (a sophisticated driving technique where the driver steers in the opposite direction of the skid) at the same time \cite{morton2006drift}. However, it is very difficult for untrained ordinary drivers to use such a specialized driving technique. As a result, oversteer often leads to severe traffic accidents; according to the National Highway Traffic Safety Administration (NHTSA) Fatality Analysis Reporting System (FARS), oversteer accounted for more than 18,852 fatal accidents in the United States from 2011 to 2020, making it the 8\textsuperscript{th} leading cause of fatal accidents. 

\begin{figure}[t!]
\centering
\subfloat[\label{fig:goal1}]{
    \includegraphics[width=0.98\linewidth]{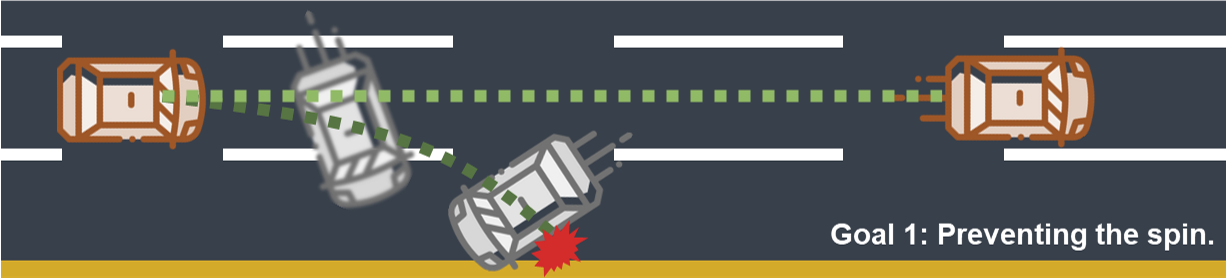}}
\hfil
\subfloat[\label{fig:goal2}]{
    \includegraphics[width=0.98\linewidth]{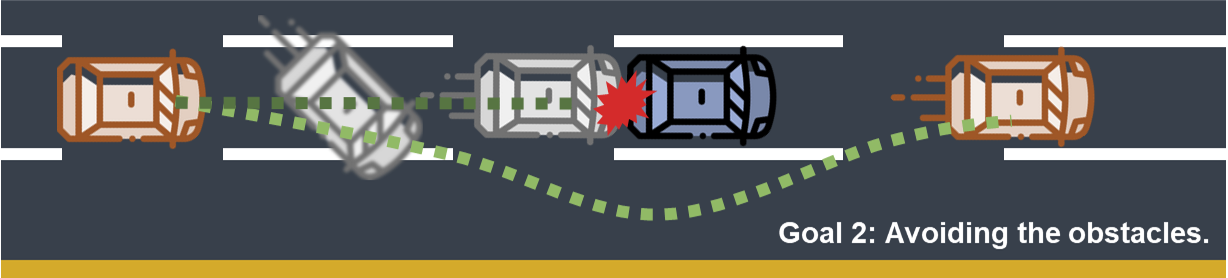}}
\caption{Research goals. (a) The ego vehicle in brown must control the oversteer in order not to spin. (b) It should also avoid the obstacle (i.e., front vehicle) in blue.}
\label{fig:goal1+goal2}
\end{figure}

To mitigate oversteer, various methods have been proposed over the years. One of the earliest and most prevalent solutions is the implementation of Electronic Stability Control (ESC), which is now standard equipment in most modern vehicles. ESC helps prevent vehicle skidding by braking individual wheels or cutting engine power. However, it can only lower the slip probability but not eliminate it entirely \cite{erke2008effects}. Moreover, since ESC adjusts wheel forces rather than steering inputs, it lacks the ability to return a vehicle that has already entered an oversteer condition back to a stable, grip state. To address the limitations of ESC, researchers have explored autonomous driving technologies that enable automatic counter-steering, allowing vehicles to correct oversteer without driver intervention. Most of these technologies focus on control techniques designed to track predefined paths even under oversteer conditions \cite{goh2018controller, velenis2011steady, zhang2018drift, zubov2018autonomous,  acosta2018teaching, cai2020high, cutler2016autonomous}. While such methods can help an oversteering vehicle return to its original lane as shown in Fig.~\ref{fig:goal1}, they may not be sufficient in real-world scenarios where surrounding obstacles (e.g., other vehicles) are present. In these situations, merely controlling the oversteer is not enough to avoid collisions. When there is an obstacle in the driving lane, as illustrated in Fig.~\ref{fig:goal2}, the vehicle must not only regain stability but also recognize the surrounding environment and execute an evasive maneuver that ensures collision avoidance and is feasible for an oversteering vehicle to perform. To tackle this challenge, our study adopts an end-to-end (E2E) approach that directly maps the vehicle's state and surrounding environment to control actions by integrating path planning and vehicle control into a single neural network. This means that the path planning process simultaneously considers the vehicle's controllability, enabling the generation of paths that are both safe and executable under oversteer conditions. 

In the realm of E2E autonomous driving, imitation learning (IL) and reinforcement learning (RL) have emerged as prominent techniques for training neural networks to perform complex driving tasks \cite{chib2023recent, le2022survey, zhu2021survey, kiran2021deep, wurman2022outracing, kaufmann2023champion}. IL involves learning control policies by mimicking expert demonstrations, and RL enables an agent to learn optimal policies through trial-and-error interactions with the environment, guided by a reward function. However, both IL and RL have fundamental limitations. The action policies trained via IL often face significant performance degradation when they encounter novel situations where expert demonstration data is scarce or when the quality of the demonstration data is poor. On the other hand, even though RL requires sufficient experience of successful episodes, as the task becomes more complex, the state and action spaces become broader, and the transition dynamics have higher variance, the probability of experiencing successful episodes through random actions diminishes. Consequently, developing an optimal action policy using RL becomes exceedingly difficult \cite{huang2023novel, zhao2021empirical}. To overcome these shortcomings, recent hybrid learning (HL) techniques combine IL and RL, utilizing expert demonstration data with RL for more efficient and improved policy development \cite{hester2018deep, rajeswaran2017learning, alakuijala2021residual, tian2021learning, lu2023imitation, gao2018reinforcement, peng2021amp}. However, HL techniques have a limitation in that the expert demonstrations must be nearly optimal or, even if noisy, must contain optimal demonstration data \cite{gao2018reinforcement}. 

Such limitations in IL, RL, and HL become more critical when it comes to the vehicle oversteer situation. Firstly, it is nearly impossible under the oversteer situation to collect optimal demonstration data, because even human drivers find it difficult to react appropriately and try immature vehicle control. As a result, the driving policy trained via IL or HL techniques could perform poorly. Moreover, while controlling the oversteer and avoiding nearby obstacles, there is an extensive amount of information to be perceived regarding the surroundings and the vehicle state, and there is a high transition variance due to the nonlinear dynamics. Not only does this make it challenging to develop an optimal driving policy using RL alone, but it also means that agents are unlikely to encounter the same situation multiple times, making it difficult to learn through repeated experiences. Therefore, there is a critical need for methods that enable fast learning from new situations and can rapidly incorporate new successful experiences into the learning process.

To overcome the aforementioned limitations, we propose an innovative HL technique, Q-Compared Soft Actor-Critic (QC-SAC), capable of learning optimal policies even from immature demonstration data and adapting swiftly to novel situations. Using the proposed QC-SAC, we trained an autonomous driving agent capable of oversteer control and collision avoidance. In scenarios where even human drivers often fail to control the vehicle, a test vehicle equipped with the driving policy developed by QC-SAC successfully avoids obstacles when it suddenly starts to spin (oversteer) on a slippery road.

The major contributions of the paper are as follows:
\begin{itemize}
    \item We introduce, to the best of our knowledge, the first safe driving technology capable of successful autonomous control of vehicles that must avoid obstacles ahead while in an oversteer condition.
    \item We propose the first benchmark for training and evaluating autonomous driving agents capable of oversteer control and collision avoidance, inspired by the actual driver training process.
    \item We present a novel HL technique, QC-SAC, that effectively utilizes immature demonstration data and adapts swiftly to novel situations. We also demonstrate the superior performance of QC-SAC in the proposed benchmark through both software-only simulation and hardware-in-the-loop simulation (HILS) validation.
\end{itemize}

The remainder of this paper is organized as follows. Section \ref{sec:prior_works} reviews prior works on autonomous oversteer control, and the representative learning techniques of E2E autonomous driving, which are IL, RL, and HL. Section \ref{sec:QC-SAC} introduces the proposed QC-SAC algorithm and Section \ref{sec:scenario_setup} describes the benchmark setup, and the state space, action space, and reward function used for the training of QC-SAC. Section \ref{sec:results} presents the experimental results, including the learning-based baseline comparison, ablation study, and HILS validation. Finally, Section \ref{sec:conclusion} concludes the paper.

\section{Prior Works} \label{sec:prior_works}

\subsection{Autonomous Oversteer Control} \label{sec:prior_works:oversteer_control}
Oversteer is a hazardous condition that poses significant challenges for vehicle stability control. Early approaches, such as Electronic Stability Control (ESC), aimed to prevent oversteer by modulating wheel forces (e.g., through selective braking or engine power reduction). While ESC effectively reduces the likelihood of slip, its reliance on wheel-force adjustments alone without direct steering interventions limits its ability to re-stabilize a vehicle already in an oversteer state. With the development of advanced steering actuation technologies, such as electronic power steering (EPS) and active front steering (AFS), it has become feasible to autonomously apply corrective steering inputs. This capability has prompted research into autonomous counter-steering methods specifically designed to address oversteer. Several studies have attempted oversteer control based on vehicle dynamics modeling \cite{goh2018controller, velenis2011steady, zhang2018drift, zubov2018autonomous, weng2024aggressive}. While vehicle dynamics models can be effective, they require extensive parameter measurements beforehand and are susceptible to model mismatches due to various external factors. For instance, the widely used Pacejka Magic Formula \cite{pacejka1992magic}, which models the nonlinear relationship between slip angle and tire force, relies on four parameters that must be measured in advance. These parameters are highly sensitive to external conditions like road surfaces, weather, tire temperature, and air pressure \cite{braghin2006environmental}, making accurate modeling challenging. Acknowledging the limitations of dynamics-based methods, learning-based control approaches have been explored. Acosta \textit{et al.} \cite{acosta2018teaching} proposed using neural networks to learn tire parameters for model-based controllers, while Cai \textit{et al.} \cite{cai2020high} and Cutler \textit{et al.} \cite{cutler2016autonomous} introduced reinforcement learning methods for drift control. 
However, both dynamics-based and learning-based methods have primarily targeted race-oriented scenarios, such as high-speed cornering or drift demonstrations, rather than focusing on traffic safety. As a result, these technologies often concentrate on tracking predefined trajectories or executing drifting maneuvers, while neglecting comprehensive path planning that includes obstacle avoidance. To our knowledge, all prior works either follow predefined expert trajectories \cite{goh2018controller, velenis2011steady, zhang2018drift, zubov2018autonomous,  acosta2018teaching, cai2020high, cutler2016autonomous} or perform simple path optimization in obstacle-free environments \cite{zhang2018drift, weng2024aggressive}.

However, on real roads with surrounding obstacles, simply following a predefined expert trajectory or conducting path optimization in an obstacle-free scenario is insufficient for safe evasive maneuvers under oversteer conditions. An appropriate evasive path that can guarantee the safety and control feasibility is required. However, without an accurate vehicle model due to the uncertainty in the slip condition as mentioned earlier, it is extremely difficult to predict the future states and optimize the trajectory. If the planned path is not dynamically feasible, the vehicle may fail to avoid obstacles or even exacerbate the unstable behavior (e.g., reverse-steer\footnote{Over-correction causing the vehicle to skid more severely in the opposite direction.}) and increase the risk of collisions. These limitations highlight the necessity of tightly integrating path planning and control. Rather than decoupling these tasks, first planning a route and then attempting to trace it, our study embraces an E2E approach. By mapping directly from the vehicle’s current state and environmental observations to control actions, the E2E approach naturally ensures that generated paths are not only safe but also feasible under oversteer conditions. This approach alleviates the dependence on highly accurate predictive models, mitigates the risk of executing infeasible maneuvers, and provides a unified solution that simultaneously considers vehicle dynamics, environmental constraints, and the need for evasive, stability-preserving trajectories.

\subsection{Imitation Learning (IL), Reinforcement Learning (RL), and Hybrid Learning (HL)} \label{sec:prior_works:IL_RL_HL}

In the field of E2E autonomous driving, two major methodologies have been widely explored: IL and RL \cite{chib2023recent, le2022survey, zhu2021survey, kiran2021deep}. In addition, hybrid learning (HL) techniques that combine the advantages of IL and RL have recently been explored. In this section, we introduce the fundamental concepts of IL, RL, and HL.

Firstly, IL aims to imitate the actions of an expert. The standard supervised imitation is also commonly referred to as behavior cloning (BC). The IL objective is generally expressed as follows:

\begin{equation}
\label{eq:ILobjective}
J_{\mathrm{IL}}(\phi;\mathcal{D})
:= -\,\mathbb{E}_{(s_d,a_d)\sim\mathcal{D}}
\left[\mathcal{L}\big(\pi_\phi(s_d),a_d\big)\right],
\end{equation}
where $\mathcal{D}$ is the dataset collected from the expert demonstration, $s_d$ is the state experienced in $\mathcal{D}$, $a_d$ is the action demonstrated by an expert in $s_d$, $\pi_\phi$ is the action policy parameterized by $\phi$ that we are training, and $\mathcal{L}$ is the loss function. The IL policy is trained to maximize $J_{\mathrm{IL}}(\phi;\mathcal{D})$, i.e., to increase the likelihood of the expert actions $a_d$ given the corresponding states $s_d$.
However, due to the nature of supervised learning, $\pi_\phi$ can output expert-like actions only for the states contained in the dataset or for states that are similar to them. Therefore, when there is a shortage of expert data, or the data is immature, significant performance degradation is inevitable.

In contrast, RL is a technique where the agent interacts with the environment to develop an action policy that can maximize the cumulative rewards. It figures out appropriate actions to maximize the cumulative rewards on each state, based on its own experiences evaluated with the predefined reward function. The objective of RL is generally formulated as follows:

\begin{equation}
\label{eq:RLobjective}
J_{\mathrm{RL}}(\phi)
:= \mathbb{E}_{(s_t,a_t)\sim\rho_{\pi_\phi}}
\left[\sum_{t=0}^{\infty}\gamma^t\, r\big(s_t,a_t\big)\right],
\end{equation}
where $\rho_{\pi_\phi}$ is the distribution of trajectories experienced by the agent using $\pi_\phi$, $r\left(s_t,a_t\right)$ is the reward at time-step $t$, and $\gamma$ is the discount factor. 

Consequently, RL is trained to maximize the expectation of cumulative rewards obtained along an episode. Because RL experiences various episodes through extensive trial and error, it has the advantage of being robust in various situations compared to IL. However, as it needs to be trained through numerous interactions with random actions, it has low sample efficiency and slow training speed. Especially, the more complex the task, the broader the state space and action space, and the higher the transition variance, the more these problems are exacerbated, making it difficult to develop the optimal action policy or even initiate the training.

To complement the weaknesses of both IL and RL mentioned above, recent research on HL has focused on combining the two techniques. From a training perspective, existing HL methods can be broadly divided into two categories: (i) methods that train IL and RL separately in stages, and (ii) methods that train IL and RL simultaneously in a single learning process.

The first category adopts a pretraining and fine-tuning paradigm, in which IL and RL are applied in successive stages. IL is first used only to initialize the policy parameters and is then discarded during subsequent RL optimization \cite{hester2018deep, rajeswaran2017learning, tian2021learning, cai2021vision, wang2019improved}. This staged approach avoids explicit gradient interference between IL and RL in a single update step, because IL and RL are not active at the same time. However, when the demonstrations are immature or suboptimal, the action policy developed through IL can diverge significantly from the optimal policy pursued by RL. Consequently, the subsequent RL phase must first undo these suboptimal behaviors before it can improve the expected return, which induces large and unstable parameter updates and substantially diminishes the effectiveness of IL pretraining.

In contrast, the second category performs simultaneous training, optimizing IL and RL jointly within a single learning process. Representative methods in this category either combine the IL and RL objectives directly \cite{lu2023imitation, goecks2020integrating} or inject an IL-related term into the reward via shaping \cite{peng2021amp, escontrela2022adversarial, brys2015reinforcement, hussein2017deep}. 
A typical example of the former is Behavior Cloned Soft Actor-Critic (BC-SAC) \cite{lu2023imitation}, whose overall training objective can be written as
\begin{equation}
    J_{\mathrm{HL}}(\phi;\mathcal{D})
    = J_{\mathrm{RL}}(\phi)
    + \lambda\, J_{\mathrm{IL}}(\phi;\mathcal{D}),
    \label{eq:hl_objective}
\end{equation}
where $J_{\mathrm{HL}}(\phi;\mathcal{D})$ denotes the hybrid learning objective and $\lambda>0$ is a weighting coefficient that balances the RL and IL objectives. 

Reward-shaping methods, on the other hand, modify the task reward by adding an imitation-related component derived from the demonstrations. 
In such approaches, the policy still maximizes an RL objective, but under a shaped reward of the form
\begin{equation}
    r'(s,a,s')
    = r_{\mathrm{task}}(s,a,s')
    + \lambda\, r_{\mathrm{IL}}(s,a,s';\mathcal{D}),
    \label{eq:hl_reward_shaping}
\end{equation}
where $r_{\mathrm{task}}(s,a,s')$ is the original task reward, $r_{\mathrm{IL}}(s,a,s';\mathcal{D})$ is an IL-related bonus that becomes larger when the policy’s behavior is closer to the demonstrations, and $\lambda>0$ again controls the strength of the imitation signal. 
Adversarial motion-prior approaches \cite{peng2021amp, escontrela2022adversarial} are representative examples, in which $r_{\mathrm{IL}}$ is typically obtained from a discriminator trained to distinguish demonstration trajectories from policy-generated trajectories.
Although these methods differ in whether the IL signal appears as an additional objective or as a shaped reward, they share the same underlying structure: at each update step, the policy is encouraged both to maximize the expected task return and to stay close to the empirical state–action distribution of the demonstrations.

\begin{figure}[t!]
    \centering
    \subfloat[\label{fig:perfectdemo}]{
    \includegraphics[width=0.48\linewidth]{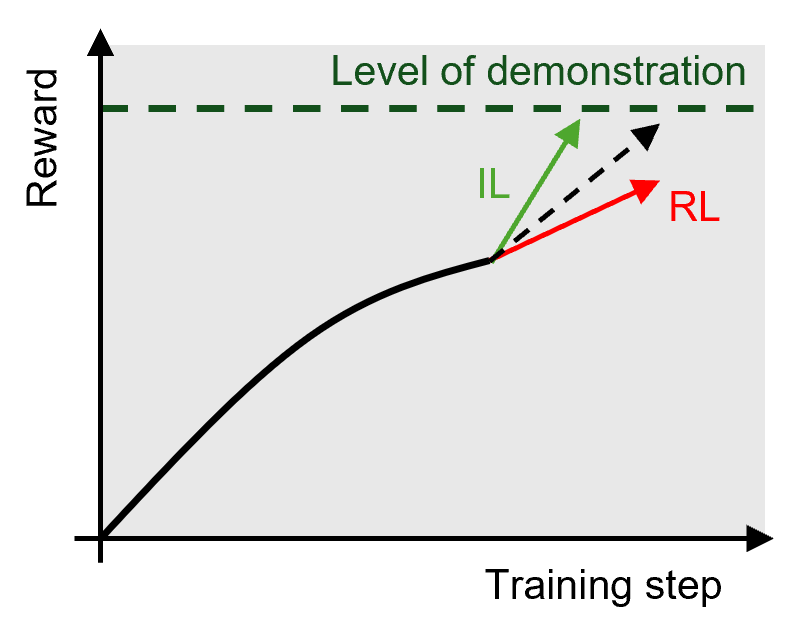}}
    \subfloat[\label{fig:poordemo}]{
    \includegraphics[width=0.48\linewidth]{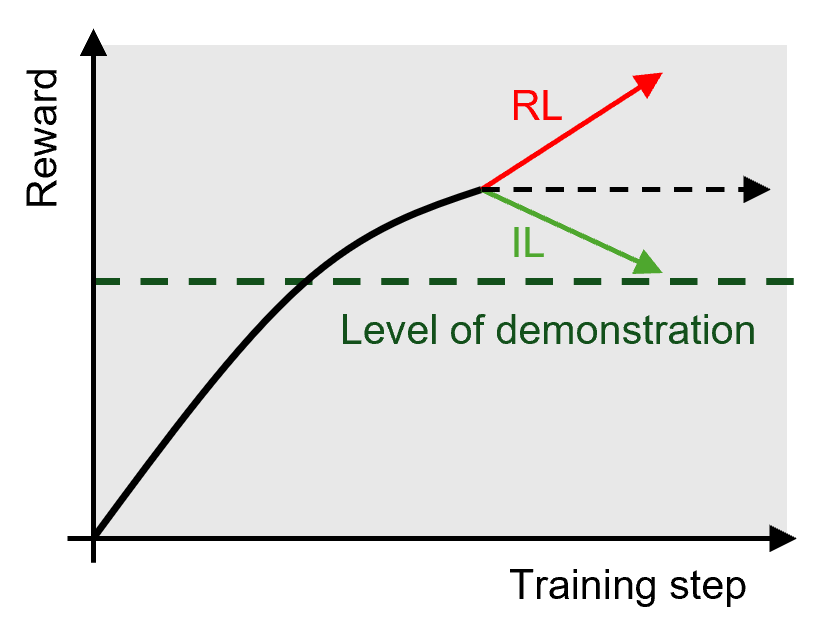}}
    \caption{Concept diagram. Impact of the quality of demonstration data on the training of action policy using existing HL techniques. (a) IL guides the policy toward the optimal region, when optimal demonstration data is given. (b) IL deteriorates the training, when immature demonstration data is given.}
    \label{fig:perfectandpoordemo}
\end{figure}

These HL methods are effective when the demonstrations are optimal, as imitation is then approximately aligned with return maximization, but they can become problematic under immature demonstrations, where the IL and RL terms may induce conflicting update directions. For clarity of exposition, we present the argument in a discrete-action setting; the same intuition extends to continuous actions by considering local probability mass shifts in action space.
Starting from the RL objective \eqref{eq:RLobjective}, we analyze its local contribution at a fixed state $s$. Conditioning on $s$ and using the action value function $Q(s,a)$, this local objective can be written as
\begin{equation}
    J_{\mathrm{RL}}^{(s)}(\phi)
    := \sum_{a} \pi_\phi(a|s)\, Q(s,a).
    \label{eq:jrl_local}
\end{equation}
Let $a_d$ be the demonstrator’s action at $s$, and let $a^* \in \arg\max_a Q^{*}(s,a)$ be an optimal action under the optimal action value function $Q^{*}$.
We consider the case in which the demonstrator is strictly suboptimal, i.e., $Q^{*}(s,a_d) < Q^{*}(s,a^*)$, and assume that the critic is locally consistent so that $Q(s,a_d) < Q(s,a^*)$.
Consider a small $\epsilon>0$ and a perturbed policy $\pi'$ defined by
$\pi'(a_d|s) = \pi_\phi(a_d|s) - \epsilon$,
$\pi'(a^*|s) = \pi_\phi(a^*|s) + \epsilon$,
and $\pi'(a|s) = \pi_\phi(a|s)$ for $a \notin \{a_d,a^*\}$.
Substituting into \eqref{eq:jrl_local} yields
\(
    J_{\mathrm{RL}}^{(s)}(\pi') - J_{\mathrm{RL}}^{(s)}(\pi_\phi)
    = \epsilon\big(Q(s,a^*) - Q(s,a_d)\big) \;>\; 0,
\)
so RL can increase its return by decreasing $\pi_\phi(a_d|s)$ and increasing $\pi_\phi(a^*|s)$.
On the other hand, when maximizing the IL objective in \eqref{eq:ILobjective}, the gradient increases $\pi_\phi(a_d|s)$ at demonstration states.
Thus, at states where $Q^{*}(s,a_d) < Q^{*}(s,a^*)$, RL improves its objective by reducing $\pi_\phi(a_d|s)$, whereas IL prefers to increase $\pi_\phi(a_d|s)$.
For objective-level hybrids of the form \eqref{eq:hl_objective}, optimizing $J_{\mathrm{HL}}(\phi;\mathcal{D})$ therefore requires reconciling two objectives that induce opposite changes in $(\pi_\phi(a^*|s),\pi_\phi(a_d|s))$ under immature demonstrations.
Reward-shaping methods of the form \eqref{eq:hl_reward_shaping} exhibit the same structural issue.
The imitation term $r_{\mathrm{IL}}(s,a,s';\mathcal{D})$ assigns higher values to actions that resemble the demonstrations even when they are immature, so the shaped reward \eqref{eq:hl_reward_shaping} biases the RL update toward actions such as $a_d$ that match the demonstrations, even at states where the task-optimal action is $a^*$.
Consequently, when the demonstrations are immature, HL methods must trade off between improving the task return and imitating suboptimal behaviors, leading to the degradation as in Fig.~\ref{fig:perfectandpoordemo}.

These limitations of IL, RL, and HL become particularly critical in the oversteer control and collision avoidance task. Firstly, it is nearly impossible to collect optimal demonstration data, because human drivers find it difficult to react appropriately and try immature vehicle control. As a result, the driving policy trained via IL or HL techniques could perform poorly. Moreover, there is an extensive amount of information to be perceived regarding the surroundings and the vehicle state, and there is a high transition variance due to the nonlinear dynamics. Not only does this make it challenging to develop an optimal driving policy using RL alone, but it also means that agents are unlikely to encounter the same situation multiple times, making it difficult to learn through repeated experiences. Therefore, there is a critical need for methods that enable fast learning from new situations and can rapidly incorporate new successful experiences into the learning process.

\section{Q-Compared Soft Actor-Critic (QC-SAC)} \label{sec:QC-SAC}

To address the aforementioned technical limitations of existing IL, RL, and HL techniques which are critical in developing an optimal driving policy capable of oversteer control and collision avoidance, we propose a novel HL technique called Q-Compared Soft Actor-Critic (QC-SAC). In this section, we introduce the three core elements of QC-SAC: Q-Compared Objective (QCO), Q-Network from Demonstration (QNfD), and Selective Demonstration Data Update (SDDU). QCO and QNfD contribute to the effective utilization of immature demonstration data to develop an optimal driving policy, while QNfD and SDDU facilitate the rapid incorporation of new successful experiences into the learning process.

\subsection{Q-Compared Objective (QCO)}
To effectively utilize immature demonstrations and support RL for optimal policy development with IL, QCO selectively utilizes only the beneficial demonstration data for IL. It estimates the action value (Q) difference between the actions stored in the demonstration data and the actions generated by the driving policy being developed. Then, it prioritizes the demonstration data that have higher Q values than the driving policy being developed, where the prioritization is implemented by weighting the demonstration data in proportion to the quality difference. QCO thus resolves a critical problem in the conventional HL techniques; the action policy becomes suboptimal when immature demonstration data are provided. 

QCO follows the general structure of the HL objective in \eqref{eq:hl_objective}, but in QC-SAC we instantiate the RL component with the Soft Actor-Critic (SAC) algorithm \cite{haarnoja2018soft, haarnoja2019soft} and the BC component with a Q-weighted behavior cloning objective. SAC is a widely used off-policy RL method for continuous control, known for its sample efficiency and training stability due to entropy regularization, and it is also adopted as the RL backbone in prior HL methods such as BC-SAC \cite{lu2023imitation}. Concretely, the overall actor objective optimized by QC-SAC is defined as
\begin{equation}
    J_{\pi}(\phi;\mathcal{D})
    = J_{\mathrm{SAC}}(\phi)
    + \lambda\, J_{\mathrm{BC}}(\phi;\mathcal{D}),
    \label{eq:qcsacobjective}
\end{equation}
where \(J_{\mathrm{SAC}}(\phi)\) and \(J_{\mathrm{BC}}(\phi;\mathcal{D})\) play the roles of \(J_{\mathrm{RL}}(\phi)\) and \(J_{\mathrm{IL}}(\phi;\mathcal{D})\) in \eqref{eq:hl_objective}, respectively. In all experiments, the weighting coefficient was fixed at \(\lambda = 1\) without task-specific tuning.

The SAC component is given by
\begin{equation} \label{eq:jsac}
    J_{\mathrm{SAC}}(\phi)
    := \mathbb{E}_{(s,a)\sim\rho_{\pi_\phi}}
    \left[Q\left(s,a\right)
    + \alpha\mathcal{H}\left(\pi_\phi\left(\cdot\middle| s\right)\right)\right],
\end{equation}
where $\mathcal{H}$ denotes the policy entropy, and $\alpha>0$ is the temperature parameter that balances exploitation and exploration.
The BC component is defined as
\begin{equation} \label{eq:jbc}
    J_{\mathrm{BC}}(\phi;\mathcal{D})
    := -\,\mathbb{E}_{(s_d,a_d)\sim\mathcal{D}}
    \left[
        C(s_d,a_d)\cdot
        \mathcal{L}_1 \big(a\sim\pi_\phi\left(s_d\right),\, a_d\big)
    \right],
\end{equation}
where $\mathcal{L}_1$ represents the L\textsubscript{1} loss, and $C(s_d,a_d)$ is the Q-compared weighting factor for each demonstration data, which is defined as
\begin{equation} \label{eq:c}
    C(s_d,a_d)
    = \max\left(
        Q^-\left(s_d,a_d\right)
        - Q\left(s_d,a\sim\pi_\phi\left(s_d\right)\right),
        0\right),
\end{equation}
where $Q^-$ is the Q-target value. 
    
In QC-SAC, the policy parameters $\phi$ are updated so as to maximize $J_{\pi}(\phi;\mathcal{D})$. The magnitude of $C(s_d,a_d)$ grows linearly with the positive gap $Q^-(s_d,a_d) - Q(s_d,a\sim\pi_\phi(s_d))$, so samples whose demonstration actions have much higher Q-values than the current policy action receive proportionally larger weights and exert a stronger influence on the policy update. Moreover, when $Q^-(s_d,a_d)\leq Q(s_d,a\sim\pi_\phi(s_d))$, the weight $C(s_d,a_d)$ becomes zero and the BC term is inactive at that state. Hence, at demonstration states the BC component in QC-SAC is either inactive or aligned with the RL objective, and the systematic IL–RL objective conflict that appears in conventional HL with immature demonstrations does not occur.

Additionally, to improve numerical stability, we use the L\textsubscript{1} loss in \eqref{eq:jbc}. In conventional HL methods such as BC-SAC \cite{lu2023imitation}, the BC objective is often implemented via the log-likelihood of demonstration actions. However, when \(a_d\) significantly differs from the distribution of \(\pi_\phi(s_d)\) because of immature demonstrations, \(\pi_\phi(a_d\vert s_d)\) can approach zero and the log probability can diverge. In contrast, the L\textsubscript{1} loss is bounded by a maximum value of 2 because the action space is constrained between \(-1\) and \(1\), preventing such divergence. Since QC-SAC explicitly allows immature demonstrations, the use of L\textsubscript{1} loss ensures numerical stability even when demonstration actions are far from the current policy actions. In this manner, the QCO of QC-SAC is specialized for the effective utilization of immature demonstrations.

\subsection{Q-Network from Demonstration (QNfD)}

Since Q values are used as the metric for evaluating $a_d$ and $a\sim\pi_\phi(s_d)$ in (\ref{eq:c}), the Q-network must be well-trained to accurately estimate $C(s_d,a_d)$, thus enabling QCO to achieve high performance. Particularly in the oversteer control and collision avoidance task, where agents frequently encounter novel situations and have difficulties experiencing successful episodes due to broad state and action spaces and high transition variance, it is crucial to obtain as much data as possible to train the Q-network effectively. To achieve this, we propose the QNfD method that utilizes demonstration data to enhance the Q-network training.

The concept of utilizing demonstration data for Q-network training was recently proposed in the HL technique employing IL-based pre-training followed by fine-tuning with RL \cite{wang2023efficient}. The study shows that when the RL is based on actor-critic, both networks for the actor and the critic need to be pre-trained. In our work, to fit the QC-SAC structure, which combines IL and RL into a single objective function without pre-training, we propose a method that combines two batches for the Q-network update. As expressed in lines \ref{alg:line:QNfDstart}$\sim$\ref{alg:line:QNfDend} of Algorithm \ref{alg:QC-SAC}, the Q-network is updated using the union $\mathcal{B}$ of the batch $\mathcal{B}_{RL}$ sampled from the replay buffer $\mathcal{D}_{RL}$ collected through interaction with the environment and the batch $\mathcal{B}_{BC}$ sampled from the demonstration dataset $\mathcal{D}$. For this purpose, not only the state and action but also the reward and next state are recorded when collecting the demonstration data. The Q-network is updated using the following conventional objective function used in RL:

\begin{equation}\label{eq:qupdate}
    J_Q(\theta_i) = -\mathbb{E}_{(s,a,r,s')\sim\pi}\left[\left( Q_{\theta_i}(s,a)-\hat Q(r,s') \right)^2\right],
\end{equation}
where $Q_{\theta_i}$ is parameterized by $\theta_i$, $i\in\{1,2\}$ is the index of two Q-networks for double Q-learning \cite{hasselt2010double}, and

\begin{equation} \label{eq:qtarget}
\hat Q(r,s')=r+\gamma \mathbb{E}_{a'\sim\pi}\left[Q_{\bar\theta_i}^-\left(s',a'\right)-\alpha \log{\pi_\phi\left(a'|s'\right)}\right].
\end{equation}
Enhancing the performance of the Q-network through QNfD can significantly improve the performance of QC-SAC, i.e., it can produce more accurate Q-value estimates for $C(s_d,a_d)$ and $J_{\mathrm{SAC}}(\phi)$ \eqref{eq:jsac} in QCO.

\begin{algorithm} [t!]
\caption{Q-Compared Soft Actor-Critic (QC-SAC)}
\begin{algorithmic}[1]
\Require demonstration dataset $\mathcal{D}$, average episode reward of demonstration $\bar{r}_{epi}$, discount factor $\gamma$, target update rate $\tau$, target entropy $\bar{\mathcal{H}}$, and learning rates $\lambda_Q$, $\lambda_\pi$, $\lambda_\alpha$

\State Initialize policy network $\pi_\phi$ and Q-networks $Q_{\theta_1}$, $Q_{\theta_2}$ with random weights
\State Initialize target Q-networks $Q^-_{\bar\theta_1}$, $Q^-_{\bar\theta_2}$ with weights $\bar\theta_1 \gets \theta_1$, $\bar\theta_2 \gets \theta_2$
\State Initialize entropy coefficient $\alpha$
\State Initialize replay buffers $\mathcal{D}_{RL} \gets \emptyset$, $\mathcal{D}_{epi} \gets \emptyset$
\State Initialize episode reward $r_{epi}\gets0$

\For{each iteration}
    \For{each environment step}
        \State $a\sim\pi_\phi(a|s)$ 
        \State $s'\sim p(s'|s,a)$ 
        \State $\mathcal{D}_{RL} \gets \mathcal{D}_{RL} \cup \left\{\left(s, a, r, s', d\right)\right\}$
        \State $\mathcal{D}_{epi} \gets \mathcal{D}_{epi} \cup \left\{\left(s, a, r, s', d\right)\right\}$
        \State $r_{epi} \gets r_{epi} + r$
        \If{$d=1$} \label{alg:line:SDDUstart}
            \If{$r_{epi} > \bar{r}_{epi}$}
                \State $\mathcal{D} \gets \mathcal{D} \cup \mathcal{D}_{epi}$
                \State $\bar{r}_{epi} \gets \left(\bar{r}_{epi}\cdot\left(|\mathcal{D}|-1\right)+r_{epi}\right)/|\mathcal{D}|$
            \EndIf \label{alg:line:SDDUend}
            \State $r_{epi}\gets0$
            \State $\mathcal{D}_{epi}\gets\emptyset$
        \EndIf
    \EndFor
    \For{each gradient step}
        \State Sample $\mathcal{B}_{RL}=\{(s,a,r,s',d)\}$ from $\mathcal{D}_{RL}$ \label{alg:line:QNfDstart}
        \State Sample $\mathcal{B}_{BC}=\{(s_d,a_d,r_d,s'_d,d_d)\}$ from $\mathcal{D}$
        \State Combined batch $\mathcal{B}=\mathcal{B}_{RL} \cup \mathcal{B}_{BC}$ \label{alg:line:QNfDend}
        \State $\theta_i \gets \theta_i + \lambda_Q \nabla_{\theta_i} J_Q(\theta_i)$ for $i\in\{1,2\}$ using $\mathcal{B}$ 
        \State $\phi \gets \phi + \lambda_\pi \nabla_\phi J_\pi(\phi)$ using $\mathcal{B}_{RL}, \mathcal{B}_{BC}$
        \State $\alpha \gets \alpha + \lambda_\alpha \nabla_\alpha J(\alpha, \bar{\mathcal{H}})$ using $\mathcal{B}_{RL}$
        \State $\bar\theta_i\gets \tau \theta_i + (1-\tau)\bar\theta_i$ for $i\in\{1,2\}$ using $\mathcal{B}$
    \EndFor
\EndFor
\end{algorithmic}
\label{alg:QC-SAC}
\end{algorithm}

\subsection{Selective Demonstration Data Update (SDDU)}

In the oversteer control and collision avoidance task, where the state and action spaces are broad and the transition variance is high, agents often encounter novel situations and find it difficult to learn by experiencing the same situations multiple times. To address this challenge, we propose the SDDU method, which enables rapid learning from new situations and the swift incorporation of new successful experiences into the learning process. SDDU selects successful episodes from interactions with the environment during the training process and uses aggregated data as the demonstration data for training. Before starting the training process, the average episode reward $\bar{r}_{epi}$ of the demonstration data is recorded, which can be calculated as: 

\begin{equation} \label{eq:repi}
\bar{r}_{epi} = \frac{1}{|\mathcal{D}|}\sum_{i=1}^{|\mathcal{D}|}\sum_{j=1}^{|\mathcal{D}[i]|}r_{i,j},
\end{equation}
where $|\mathcal{D}|$ is the size of the demonstration dataset $\mathcal{D}$, $|\mathcal{D}[i]|$ is the number of steps in the $i$\textsuperscript{th} episode in $\mathcal{D}$, and $r_{i,j}$ is the reward of the $j$\textsuperscript{th} step in the $i$\textsuperscript{th} episode.

Subsequently, while the agent interacts with the environment for training, at the end of each episode, the episode reward $r_{epi}$ (i.e., the sum of reward received in that episode) is compared to $\bar{r}_{epi}$ of the dataset. If the reward $r_{epi}$ of the new episode is higher than $\bar{r}_{epi}$, the episode is added to the dataset, and $\bar{r}_{epi}$ is updated again using (\ref{eq:repi}). This iterative process is shown in lines \ref{alg:line:SDDUstart}$\sim$\ref{alg:line:SDDUend} of Algorithm \ref{alg:QC-SAC}. By employing SDDU, the agent effectively expands its knowledge base with higher-quality data, enhancing the learning stability and performance of the driving policy. SDDU not only alleviates the data scarcity problem in IL by increasing the size and quality of the dataset, but also contributes to the effective maximization of QCO. After the iterative process of SDDU, we obtain the updated dataset $\mathcal{D}'$, which satisfies $\mathbb{E}_{(s,a) \sim \mathcal{D}'}[r(s,a)] > \mathbb{E}_{(s,a) \sim \mathcal{D}}[r(s,a)]$. Then, from $Q(s,a)=\mathbb{E}\left[\sum_{k=0}^\infty \gamma^k r_{t+k} | s_t=s, a_t=a\right]$, it follows that $\mathbb{E}_{(s,a) \sim \mathcal{D}'}[Q(s,a)] > \mathbb{E}_{(s,a) \sim \mathcal{D}}[Q(s,a)]$. Therefore, from (\ref{eq:c}), $\mathbb{E}_{(s,a) \sim \mathcal{D}'}[C(s,a)] > \mathbb{E}_{(s,a) \sim \mathcal{D}}[C(s,a)]$, and thus $J_{BC}(\phi)$ is considered more importantly, promoting active learning from expert behavior. Additionally, training $\pi_\phi$ by maximizing $J_{\mathrm{BC}}(\phi)$ with $\mathcal{D}'$ leads to higher $\mathbb{E}_{(s,a)\sim\rho_{\pi_\phi}}[Q(s,a)]$, contributing to the maximization of $J_{\mathrm{SAC}}(\phi)$ as well.

In summary, the proposed QC-SAC technique consists of three key elements: QCO, QNfD, and SDDU. The overall structure of QC-SAC can be found in Algorithm \ref{alg:QC-SAC}. We use (\ref{eq:qupdate}) to update the Q-network and (\ref{eq:qcsacobjective}) to update the $\pi$ network. The temperature $\alpha$ is updated using the same formula as in the second version of SAC \cite{haarnoja2019soft}.

\begin{equation}\label{eq:alphaupdate}
    J(\alpha, \bar{\mathcal{H}}) = \mathbb{E}_{a\sim\pi}\left[\alpha\log{\pi(a|s)}+\alpha \bar{\mathcal{H}}\right]
\end{equation}

\subsection{Focused Experience Replay (FER)}

To improve training quality, we employ Focused Experience Replay (FER) \cite{kong2021enhanced}. FER addresses the data imbalance problem in conventional random sampling, where older data in the replay buffer is more likely to be sampled than recent data. By using a half-normal distribution for sampling, FER prioritizes recent data, enhancing training speed and stability.

\section{Oversteer Control and Collision Avoidance Benchmark Setup} \label{sec:scenario_setup}

\begin{figure*}[htb!]
\centering
\begin{minipage}[b]{0.384\textwidth}
    \subfloat[\label{fig:kickplatea}]{
        \includegraphics[width=0.98\textwidth]{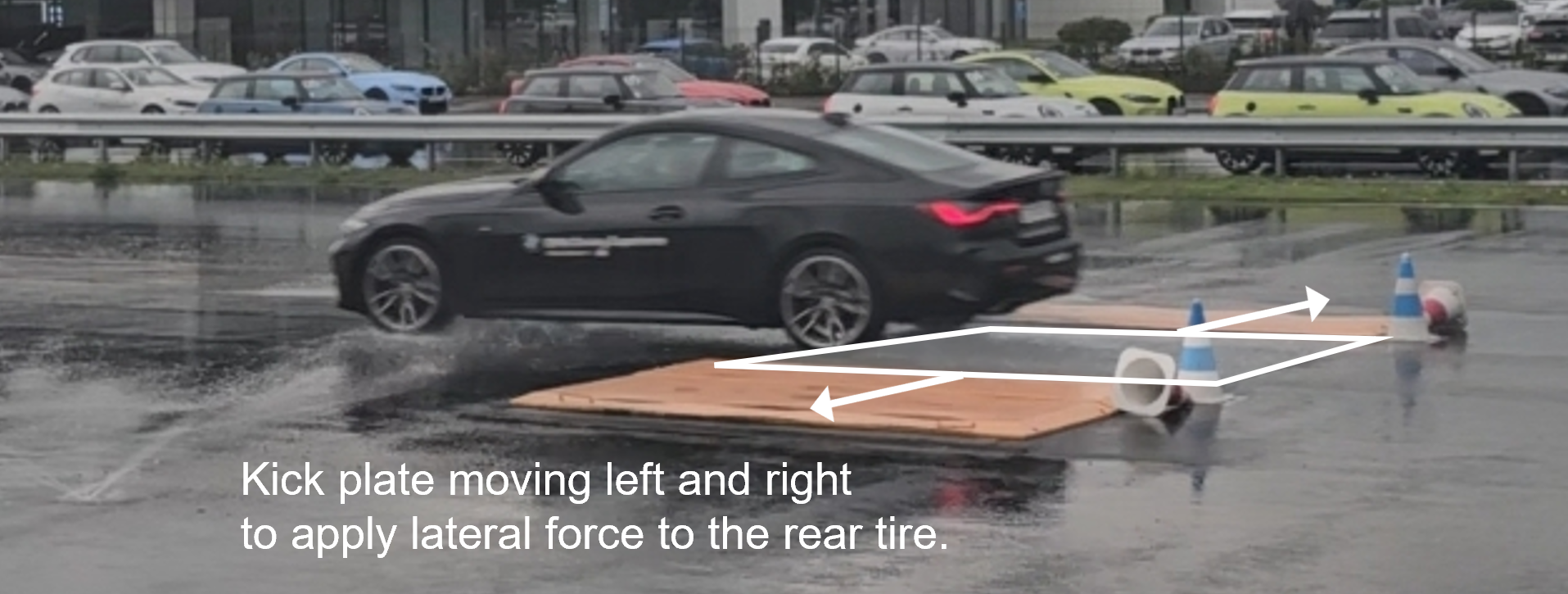}}\\
    \subfloat[\label{fig:kickplateb}]{
        \includegraphics[width=0.98\textwidth]{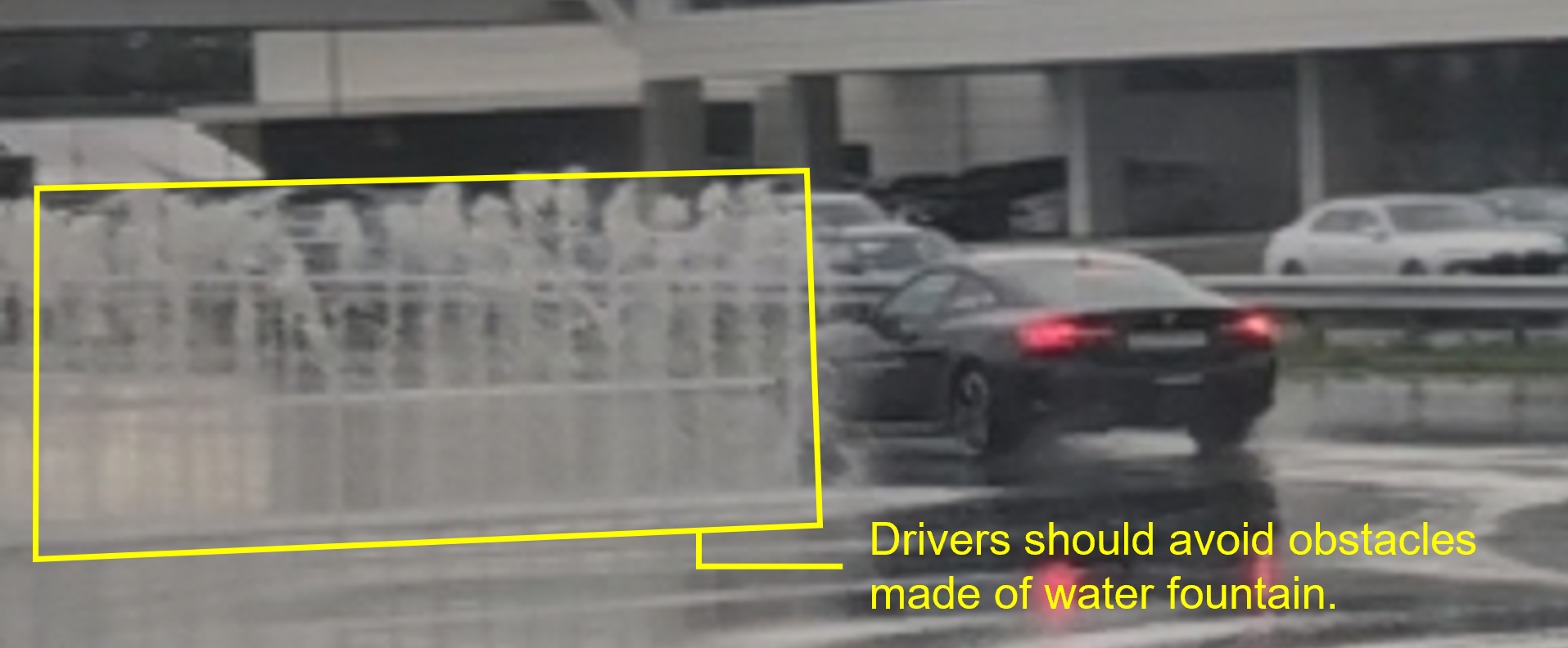}}
\end{minipage}
\begin{minipage}[b]{0.596\textwidth}
    \subfloat[\label{fig:kickplatec}]{
    \includegraphics[width=0.98\textwidth]{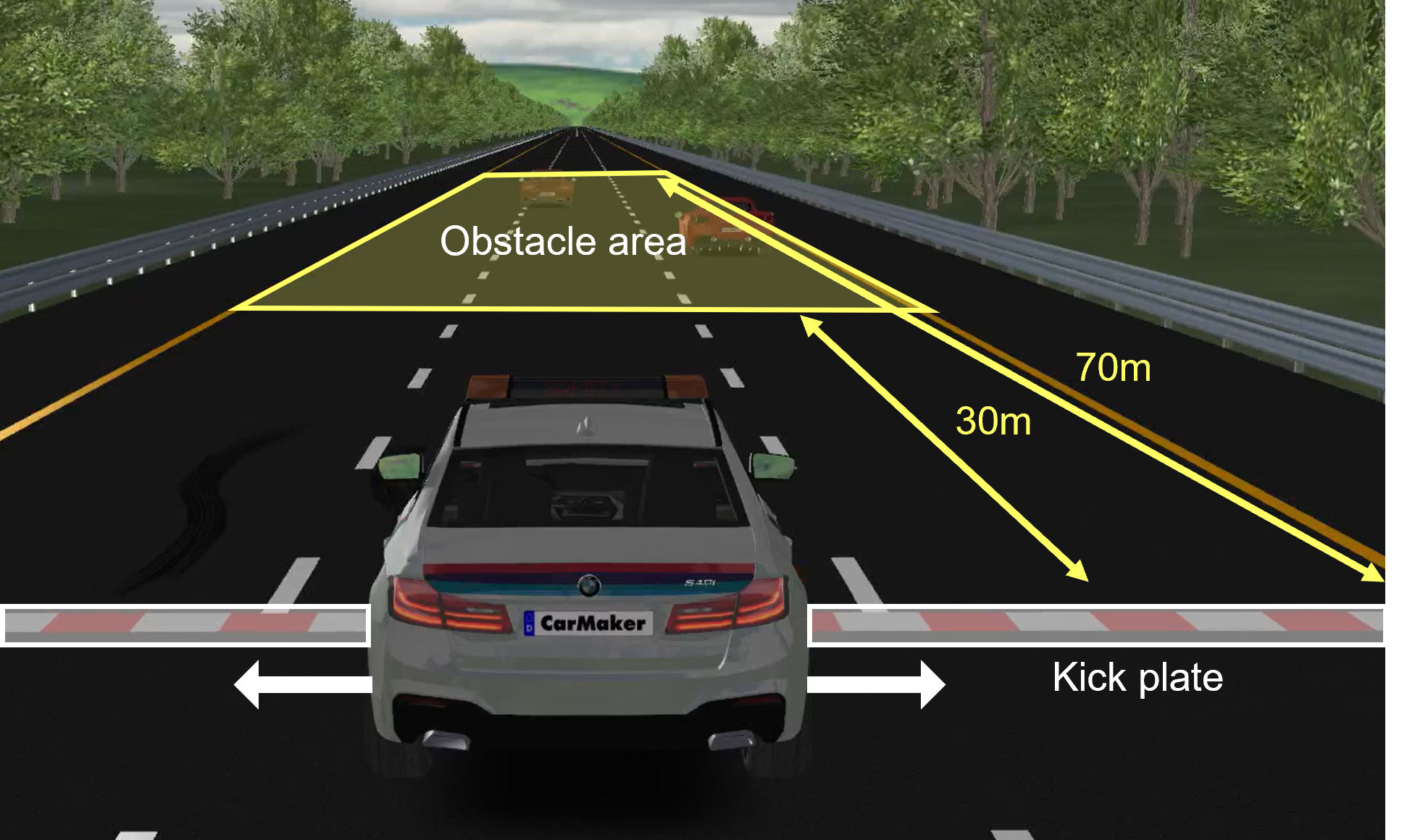}}  
\end{minipage}
\caption{Real-world driver training process and oversteer control and collision avoidance benchmark in a virtual environment. (a) Kick plate inducing oversteer, and (b) a collision avoidance scenario for driver training at BMW Driving Center, South Korea. (c) Oversteer control and collision avoidance benchmark developed in IPG CarMaker simulator.}
\label{fig:kickplate}
\end{figure*}

\subsection{Benchmark Setup} \label{sec:benchmark}

To develop the E2E driving policy network capable of oversteer control with collision avoidance, we propose a novel training and testing benchmark. The benchmark is inspired by the real driver training procedure which utilizes a kick plate: a device that induces oversteer intentionally by moving laterally when the rear wheels of the vehicle pass over it (see Fig.~\ref{fig:kickplatea}). The human driver must control the oversteer while avoiding water fountains or virtual obstacles, as illustrated in Fig.~\ref{fig:kickplateb}. For safety reasons, we implement this within a simulator, creating a virtual kick plate to induce oversteer. Obstacles are then randomly placed on the road so that the autonomous vehicle encounters them after the oversteer is induced. 

To compose the oversteer control and collision avoidance benchmark, a kick plate and obstacles are configured in the simulator. First, to deliberately induce oversteer, road friction is reduced to 40\%, and a virtual kick plate is configured to apply lateral force to the rear tires as the vehicle passes a certain point. The lateral force applied to the rear tires is randomly set to vary the intensity and direction of the oversteer. The learned controller is activated only after the kick-plate disturbance has ended; thus, the kick plate serves solely to generate the initial oversteer condition. Additionally, as shown in Fig.~\ref{fig:kickplatec}, obstacles are randomly placed within 30 to 70 m distance ahead of the kick plate. The distance between the kick plate and the obstacles is set closer than the legally recommended safe distance\footnote{Converting the distance between the kick plate and the obstacles to Time to Collision (TTC) at an entry speed of 70 km/h, 30 m corresponds to 1.54 s, and 70 m corresponds to 3.60 s, which are within the legally defined safe distances. Generally, the safety distance required is 2 s in the US and Europe and 3.6 s in South Korea.}, rendering simple braking insufficient for collision avoidance and necessitating the formulation of precise avoidance trajectories. Furthermore, to prevent the cases where the road is entirely obstructed and driving becomes unfeasible, the number of obstacles placed is one fewer than the number of lanes. In the experiments, up to two obstacles are randomly positioned on a three-lane road.

\subsection{State Space} \label{sec:state_space}

\begin{figure}[t!]
\centering
\subfloat[\label{fig:vehiclestate}]{
    \includegraphics[width=0.455\linewidth]{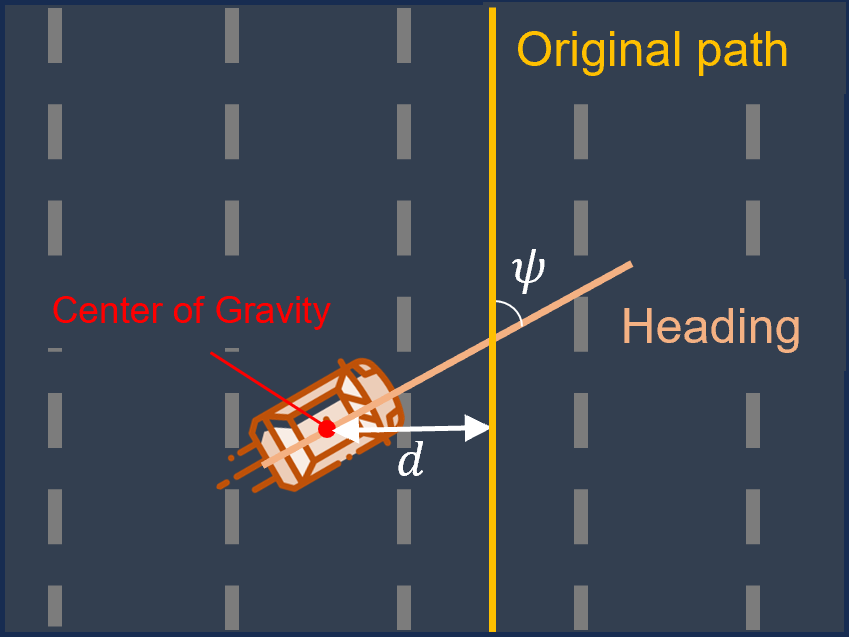}}
\subfloat[\label{fig:surroundingstate}]{
    \includegraphics[width=0.525\linewidth]{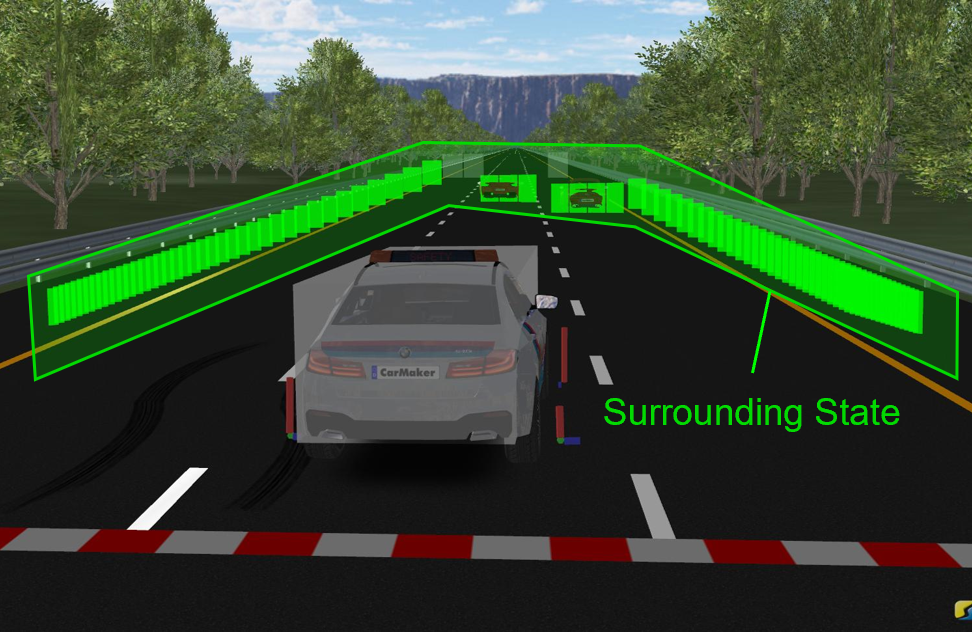}}\\
\subfloat[\label{fig:rewardfunction}]{
    \includegraphics[width=0.98\linewidth]{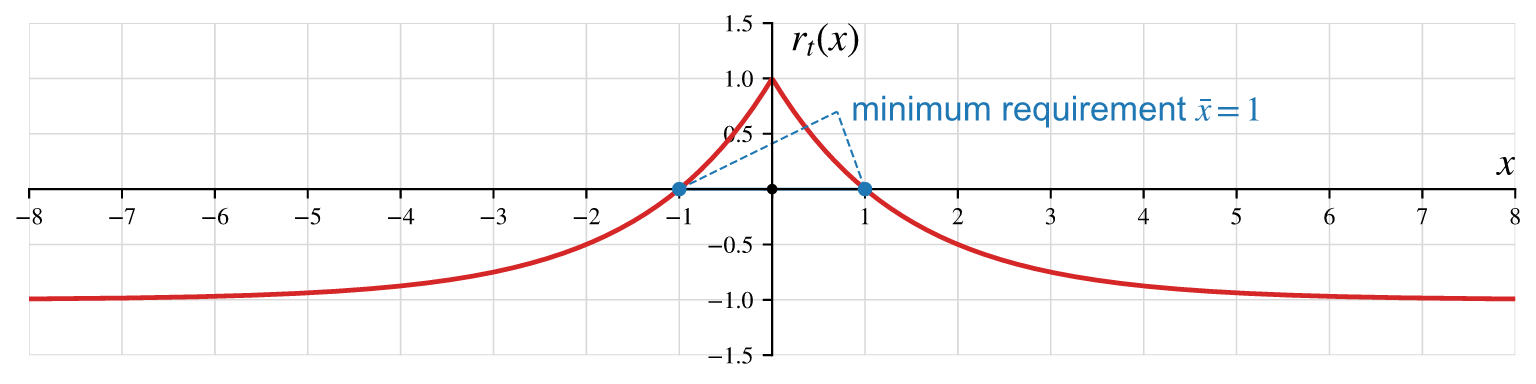}}\\
\caption{Experimental setup. (a) Definition of $d$ and $\psi$ in vehicle state. (b) Representation of surrounding state. (c) An example of the reward function. ($\bar{x}=1$).}
\label{fig:experimentalsetup}
\end{figure}

The state space serving as input for RL and IL models is divided into two categories: vehicle state and surrounding state. The vehicle state includes eight values: side slip angle ($\beta$), longitudinal velocity ($v_{long}$), longitudinal acceleration ($a_{long}$), lateral acceleration ($a_{lat}$), cross-track error from original path or lane ($d$), orientation error with respect to original path ($\psi$), yaw rate ($\dot{\psi}$), and steering angle ($\delta$). Side slip angle is calculated from $v_{long}$ and lateral velocity ($v_{lat}$) ($\beta = \arctan\left(\frac{v_{lat}}{v_{long}}\right)$), while $d$ and $\psi$ are calculated based on the center line of the lane in which the vehicle was traveling before oversteering and the vehicle’s center of gravity, respectively (see Fig.~\ref{fig:vehiclestate}). The reason for considering the steering angle in the vehicle state will be discussed in the next section, with the action space setup. The surrounding state is composed of a 90-dimensional vector. This is a vectorized drivable area within a 90\textdegree{} azimuth range centered on the front direction of the vehicle with 1\textdegree{} resolution, representing the drivable distance for each azimuth angle in polar coordinates. For each azimuth angle, the shortest distance to an obstacle or the road boundary is recorded. The visualization of the surrounding state vector can be seen in Fig.~\ref{fig:surroundingstate}.

\subsection{Action Space and Action Constraint}

The action space is represented with a two-dimensional vector of pedal value and steering angular velocity. Generally, in E2E autonomous driving systems, the output action consists of longitudinal and lateral control values. For longitudinal control, it is typically either the target speed \cite{tian2021learning} or the pedal value \cite{wurman2022outracing, cai2020high, cutler2016autonomous, wang2019improved}, while the lateral control value is usually the steering angle \cite{wurman2022outracing, tian2021learning, cai2020high, cutler2016autonomous, wang2019improved}. In the oversteer situation considered in this study, changes in weight transfer or slip ratio caused by pedal manipulation are more critical than the vehicle's speed. Therefore, the pedal value is selected as the longitudinal control value instead of the target speed. For lateral control, steering angular velocity is used instead of the steering angle to constrain the steering angular velocity. Note that when the steering angle is chosen as an action output, we cannot constrain the steering angular velocity, as the policy network can output very different steering angles at two consecutive time-steps. Indeed, when we develop a driving policy to produce the steering angle as an action, the maximum steering angular velocity records 18,000\textdegree/s. That means when the simulation runs at 20 Hz and the test vehicle has a steering range from -450\textdegree{} to 450\textdegree{}, a driving policy that steers with the maximum angular velocity is trained. An experiment video can be found at \url{https://youtu.be/BD4dNo4OM7k}.
Since it is impractical to control the steering wheel at such a high angular velocity, we propose a method to constrain the vehicle's steering angular velocity. By specifying the steering angular velocity ($\dot{\delta}$) as an action output and considering the steering angle ($\delta$) as part of the state, the steering angular velocity can be constrained without violating the Markov property. Considering the Markov property that the current state $s_{t}$ is influenced only by the previous state $s_{t-1}$ and the previous action $a_{t-1}$, and the definition that $\delta_{t-1}\in s_{t-1}$ and $\dot{\delta}_{t-1}\in a_{t-1}$, $\delta_{t} \in s_{t}$ can be expressed as $\delta_{t}=\delta_{t-1}+\dot{\delta}_{t-1}\cdot \Delta t$. Since time step $\Delta t$ is fixed at 0.05 s in our 20 Hz simulation setup, our method does not violate the Markov property. In the simulation, both outputs, pedal value and steering angular velocity, are normalized between -1 and 1. Negative and positive pedal values indicate the brake and throttle pedal manipulations, respectively. The steering angular velocity is set by multiplying the output value by 700\textdegree/s, ensuring that the maximum steering angular velocity is 700\textdegree/s. This value is based on the maximum steering angular velocity of the steer-by-wire system \cite{yih2005modification}.

\subsection{Reward Function Shaping}

The reward function consists of four components: safe distance reward ($R_{safe}$), progress reward ($R_{prog}$), auxiliary reward ($R_{aux}$), and terminal reward ($R_{term}$). Except for $R_{prog}$ and $R_{term}$, all reward functions adhere to the form of 

\begin{equation} \label{eq:reward_form}
    r_t\left(x\right)={0.5}^{\left|x\right|/\bar{x}}\times2-1,
\end{equation}
where $x$ is the value of interest, and $\bar{x}$ is the predefined requirement value for $x$. This form of reward function is suitable for situations where a smaller $x$ value is desired. It limits the range of the reward value and considers the requirement value. Starting from an exponential function $r_t\left(x\right)=e^{-k\left|x\right|}$, it is modified to $r_t\left(x\right)=e^{-k\left|x\right|}\times2-1$ to limit the reward value between -1 and 1. The x-intercept (at $x=\pm\frac{\ln{\left(2\right)}}{k}$) is set as the requirement value we aim to achieve at least. Aligning the x-intercept with the requirement value yields (\ref{eq:reward_form}). The visualization of the reward function form is shown in Fig.~\ref{fig:rewardfunction}.

Using the above function form in (\ref{eq:reward_form}), the four reward components at time-step $t$ are defined as
\begin{align}
R_{safe,t} &= -r_t\!\left(\min_{i} d^{\mathrm{sur}}_{t,i}\right),
\label{eq:reward_safe} \\
R_{prog,t} &= \ell^{\mathrm{ref}}_t - \ell^{\mathrm{ref}}_{t-1},
\label{eq:reward_prog} \\
R_{aux,t} &= r_t(d_t) + r_t(\sqrt{a_{long,t}^2 + a_{lat,t}^2}) + r_t(\beta_t) + r_t(\dot{\delta}_t),
\label{eq:reward_aux} \\
R_{term,t} &\in \{-50, 0, 50\},
\label{eq:reward_term}
\end{align}
where $d^{\mathrm{sur}}_{t,i}$ denotes the drivable distance for the $i$-th azimuth angle in the surrounding state vector (Section \ref{sec:state_space}), and $\ell^{\mathrm{ref}}_t$ is the distance traveled in the tangent direction within the Frenet–Serret frame with respect to the path planned before the oversteer. 
$R_{safe,t}$ encourages the agent to maximize the safety distance from obstacles by penalizing small values of $\min_i d^{\mathrm{sur}}_{t,i}$ in the total reward, and $R_{prog,t}$ forces the vehicle to move forward by calculating the difference in the distance traveled between time-steps $t$ and $t-1$. 
$R_{aux,t}$ aims to enhance passenger comfort, improve control stability, and find an avoidance path that does not deviate significantly from the original path (i.e., the path planned before the oversteer) whenever possible. 
Finally, $R_{term,t}$, added at the end of each episode, evaluates whether the episode was successful: when the side slip angle remains stable at less than $1^\circ$ for 100 time-steps, we consider that the vehicle is in the grip state, and the episode ends, granting a reward of $+50$; when the vehicle goes off-road, collides with an obstacle, or the side slip angle exceeds $37^\circ$ (the vehicle's maximum wheel steer angle) and spins, a penalty of $-50$ is given.

The total reward ($R_{tot}$) that combines all four reward components is defined as follows:
\begin{equation} \label{eq:reward_total_OCCA}
R_{tot,t}=\lambda_1R_{safe,t}+\lambda_2R_{prog,t}+\lambda_3R_{aux,t}+R_{term,t},
\end{equation}
where $\lambda_i$ for $i \in \{1,2,3\}$ is the weighting factor for each reward term.

In this study, the reward weights are selected according to the task priorities. Since this study addresses a safety-critical driving task, the safety-distance term is assigned two-thirds of the total weight of the three non-terminal reward terms ($\lambda_1=0.8$), while the remaining one-third is equally divided between the progress and auxiliary terms ($\lambda_2=\lambda_3=0.2$).  Additionally, the requirement values are set as follows: safe distance to 5 m, cross-track error to 3.5 m (the width of one lane), acceleration to 2.943 m/s\textsuperscript{2}, side slip angle to 20\textdegree, and steer rate to 300\textdegree/s. 

\section{Performance Evaluation} \label{sec:results}

We evaluate QC-SAC from three perspectives. First, we compare QC-SAC with representative learning-based baselines in the proposed oversteer control and collision avoidance benchmark. Second, we conduct an ablation study to examine the contribution of each component of QC-SAC. Third, we validate the trained policy in a HILS setup with physical steering hardware feedback. Additional analyses of vehicle and road parameter variations, reward-weight sensitivity, and physics-based controller comparisons are provided in the Appendix.

\subsection{Learning-based Baseline Comparison}
\label{sec:baseline_comparison}

To evaluate the performance of QC-SAC, we compare the performance of driving policy developed with QC-SAC to those of four other representative and widely used IL, RL, and HL baselines: Behavior Cloning (BC), Soft Actor-Critic (SAC) \cite{haarnoja2019soft}, Behavior Cloned Soft Actor-Critic (BC-SAC) \cite{lu2023imitation}, and Adversarial Motion Priors (AMP) \cite{peng2021amp}. In this section, we demonstrate the superiority of QC-SAC by comparing the performance of driving policies in the proposed benchmark.

\begin{figure*}[t!]
    \centering
    \subfloat[\label{fig:result_OCCA}]{
        \includegraphics[width=0.326\textwidth]{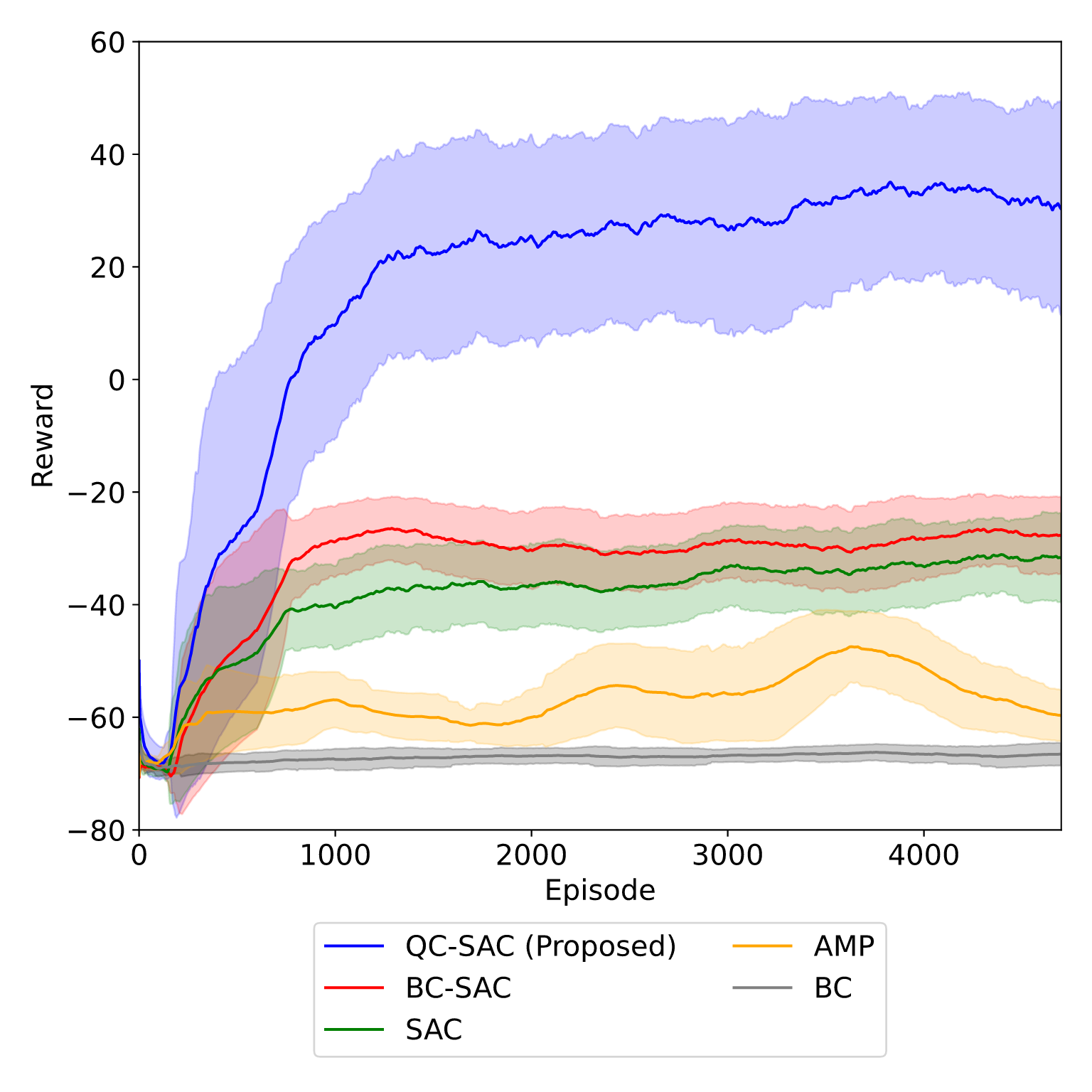}}
    \subfloat[\label{fig:ablation_a}]{
        \includegraphics[width=0.326\textwidth]{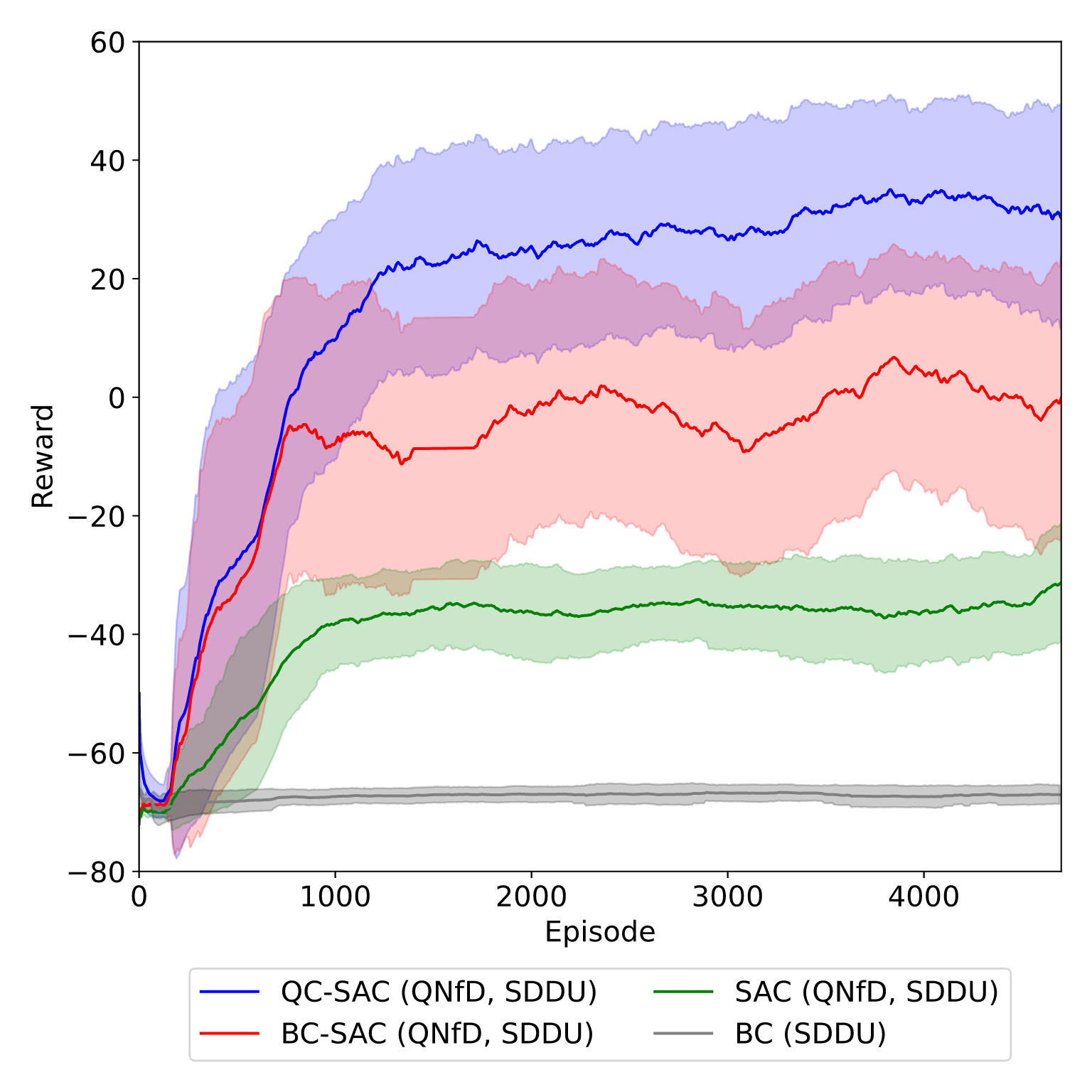}}
    \subfloat[\label{fig:ablation_b}]{
        \includegraphics[width=0.326\textwidth]{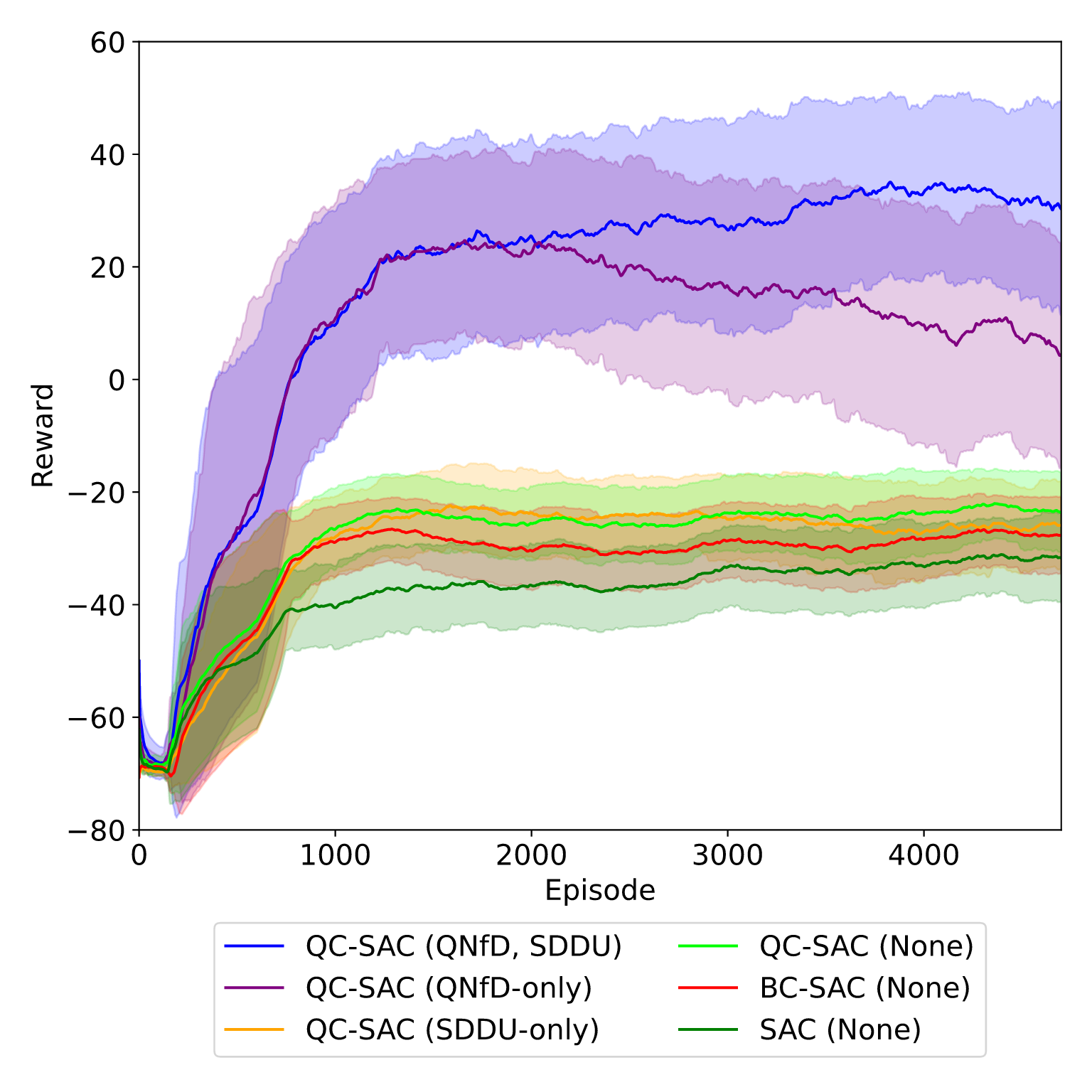}}\\
    \caption{Training curve. The solid line represents the average reward of five instances of each training technique initialized with random seeds, while the shaded area indicates their standard deviation. The proposed QC-SAC achieves significantly higher rewards compared to other techniques. (a) Performance evaluation, and (b,c) ablation study in the proposed benchmark. Because of the random obstacle placements, there are cases when the autonomous driving agent may not have any possible path to avoid the obstacles and the optimal policy inevitably experiences collisions. As a result, the standard deviation of the QC-SAC still remains high in the converged state.}
    \label{fig:trainingcurve} 
\end{figure*}

To develop a driving policy that can control oversteer and avoid nearby obstacles, we first build a dataset from demonstrations of a human driver with a racing wheel input device. A total of 25,567 time-steps, equivalent to 21.3 minutes of driving data, are collected from 200 episodes. Using the collected data, we develop driving policies using five techniques: BC, SAC, BC-SAC, AMP and QC-SAC. The resulting reward graphs per training episode are shown in Fig.~\ref{fig:result_OCCA}. Additionally, we infer the driving policies and conduct 500 episodes of test runs to compare the control \& avoidance success rates of each approach. During the test runs, the location of obstacles and the intensity and direction of the oversteer induced by the kick plate are completely randomized. The results are shown in TABLE~\ref{table:OCCA}. 

The reward graph and success rate of BC, which only learns from demonstration data without any interaction with the environment, testify the quality of the given demonstrations. As shown in Fig.~\ref{fig:result_OCCA}, it is impossible to develop a good driving policy by solely using the given immature demonstration data. SAC, which maximizes rewards through the interaction with the environment, shows higher rewards than BC. However, as it is very difficult to experience successful episodes through random actions in the oversteer control and collision avoidance task, which has broad state and action spaces and high transition variance, SAC fails to develop the optimal driving policy. Note that selecting one wrong action in oversteer situation severely destabilizes the vehicle, making it impossible to complete the episode successfully. The probability that SAC consistently outputs appropriate actions at every time-step within an episode through random exploration is extremely low. Among the HL baselines, BC-SAC, which considers both demonstration data and the interaction with the environment, performs slightly better than SAC but still cannot achieve high rewards as it continuously relies on the immature demonstration data that disturbs the training.
Also, AMP underperforms SAC in this task because the adversarial imitation reward distorts the original task reward toward immature demonstrations, corrupting the reward function and severely degrading policy learning.

\begin{figure}[t!]
\centering  
\includegraphics[width=\linewidth]{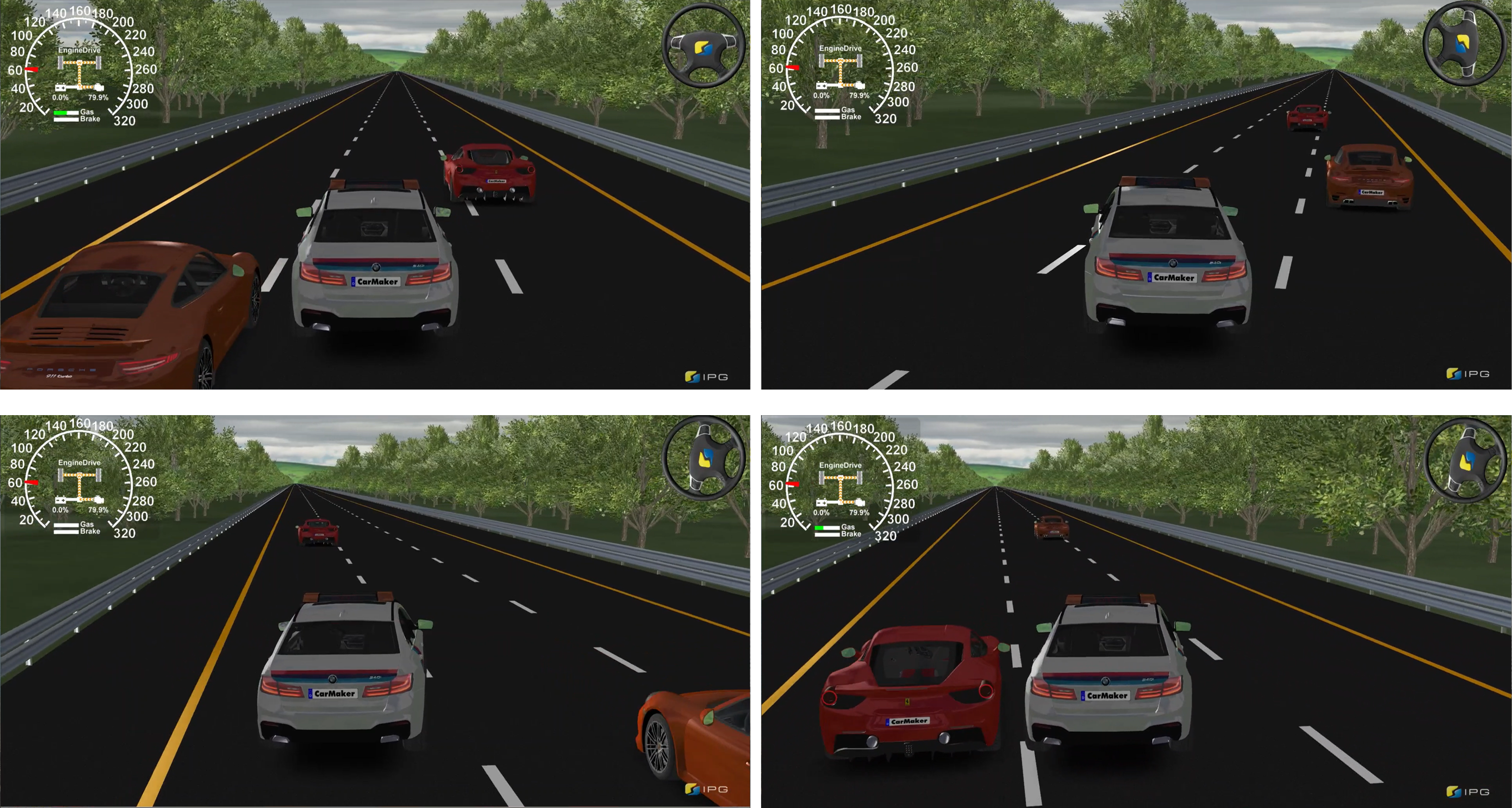}
\caption{Captured scenes during the test runs. Videos of the test runs using QC-SAC in the oversteer control and collision avoidance benchmark can be found at \url{https://youtu.be/ofiRob-hl6g}.}
\label{fig:testrun}
\end{figure}

In contrast, QC-SAC, which can effectively utilize immature demonstration data, records significantly higher rewards than other representative techniques. In the result of the test runs shown in TABLE~\ref{table:OCCA}, the proposed technique records a control \& avoidance success rate of about 81.8\%. While this success rate might seem insufficient in the autonomous driving field, where safety is critical, it represents a significant improvement given the 34.6\% success rate of existing techniques and the challenges human drivers encounter in the same situation. Indeed, the human driver who collected the dataset recorded about a 15\% success rate over 100 test runs when the positions of the front obstacles and the intensity and direction of the kick plate were unknown. Additionally, among the 91 episodes where QC-SAC fails to avoid collision, 95.6\% (87 episodes) involve obstacles completely blocking the path in the direction of the vehicle's skid caused by the kick plate, making collision avoidance physically impossible within the given action space. The occurrence of such unavoidable scenarios under randomized benchmark conditions also accounts for the relatively large standard deviation of the QC-SAC reward after convergence in Fig.~\ref{fig:result_OCCA}. Examples of these unavoidable collision failure cases can be seen in the video linked in Fig.~\ref{fig:testrun}. Therefore, excluding unavoidable collisions, the driving policy developed via QC-SAC achieves a 99.0\% success rate, demonstrating an almost optimal (i.e., near-optimal) performance.

\begin{table}[t!] 
\caption{Success rate for 500 episodes of test runs.} \label{table:OCCA}
\begin{center}
\begin{tabular}{c|ccc}
\toprule
\textbf{Method} & \textbf{Success Rate} & \textbf{Success} & \textbf{Fail} \\
\midrule
QC-SAC (Proposed)  & 81.8\%  & 409  & 91\\
BC-SAC \cite{lu2023imitation}  & 34.6\%  & 173  & 327\\
SAC \cite{haarnoja2019soft}  & 19.0\%  & 95  & 405\\
AMP \cite{peng2021amp}  & 8.6\%  & 43  & 457\\
BC  & 0.4\%  & 2  & 498\\
\bottomrule
\end{tabular}
\end{center}
\end{table}

\subsection{Ablation Study}

An ablation study is conducted on the 3 core elements of QC-SAC (i.e., QCO, QNfD, and SDDU). First, to evaluate the impact of QCO, QNfD and SDDU are applied to BC-SAC, SAC, and BC, with all other conditions kept the same except for the objective function. Note that QNfD is not applicable to BC because it does not use Q values for training. The comparison results are shown in Fig.~\ref{fig:ablation_a}. Compared to Fig.~\ref{fig:result_OCCA}, it can be seen that the reward of BC-SAC slightly improves due to the application of QNfD and SDDU, but still records significantly lower rewards compared to the proposed QC-SAC. This confirms that under immature demonstrations, QCO resolves the fundamental limitation of conventional HL methods, the IL–RL objective conflict, and is essential for learning an optimal action policy.

For the evaluation of QNfD and SDDU, we remove each function from the QC-SAC and observe the performance degradation. The results with SDDU and QNfD removed from the QC-SAC are labeled as QNfD-only and SDDU-only, respectively, while the results with both methods removed are labeled as None. As shown in Fig.~\ref{fig:ablation_b}, performance declines when SDDU and QNfD are not used; in particular, when QNfD is removed, there is a significant performance degradation. Because QC-SAC evaluates demonstration data based on Q values, when the performance of the Q-network decreases as the demonstration data is not used in Q-network training, the overall QC-SAC shows a substantial performance drop. This proves that QNfD is essential for QCO. On the other hand, when SDDU is discarded from QC-SAC, the reward graph starts to decline after around 2,000 episodes of training. This can be expected due to the overfitting from continuous behavior cloning on a fixed small set of demonstration data. Among the collected 25,567 time-steps of data, excluding those data with low Q values that are not considered by the QCO, a small amount of data could have been used for training, which causes overfitting. In contrast, QC-SAC, which continuously updates and improves the demonstration data with successful episodes, does not encounter an overfitting problem. Lastly, when both QNfD and SDDU are not implemented in QC-SAC, the performance of QC-SAC degrades significantly. However, due to the significant degradation by the removal of QNfD, the difference between None and QNfD-only is not substantial. 

The results of the ablation study prove that all three core elements of QC-SAC, QCO, QNfD, and SDDU, are essential and critical. Therefore, the best performance is achieved when all of these three elements are employed.

\subsection{Hardware-in-the-Loop Simulation Validation}
\label{sec:hils_validation}

To further address the gap between software-only simulation and real vehicle implementation, we conducted HILS validation by incorporating a physical steering hardware interface into the closed-loop evaluation. In this setup, the vehicle dynamics and surrounding environment are simulated in IPG CarMaker using the high-fidelity vehicle and tire configuration described in Appendix~\ref{app:vehicle_model_parameters}, while the lateral control command generated by QC-SAC is executed through physical steering hardware.

\begin{figure}[t!]
\centering  
\subfloat[\label{fig:hils_setup}]{
    \includegraphics[width=0.98\linewidth]{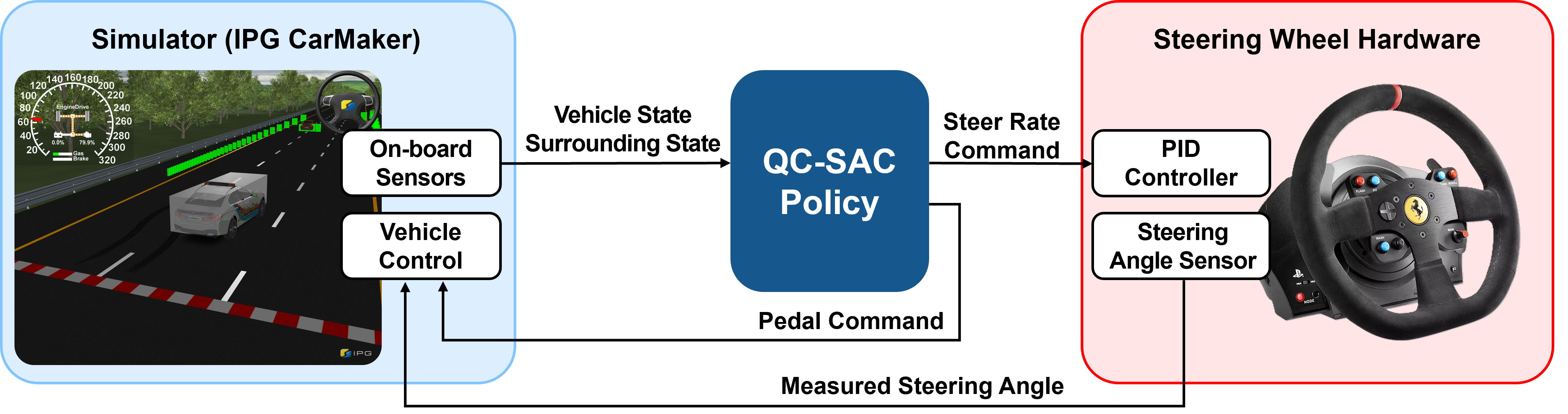}}\\
\subfloat[\label{fig:hils_capture}]{
    \includegraphics[width=0.98\linewidth]{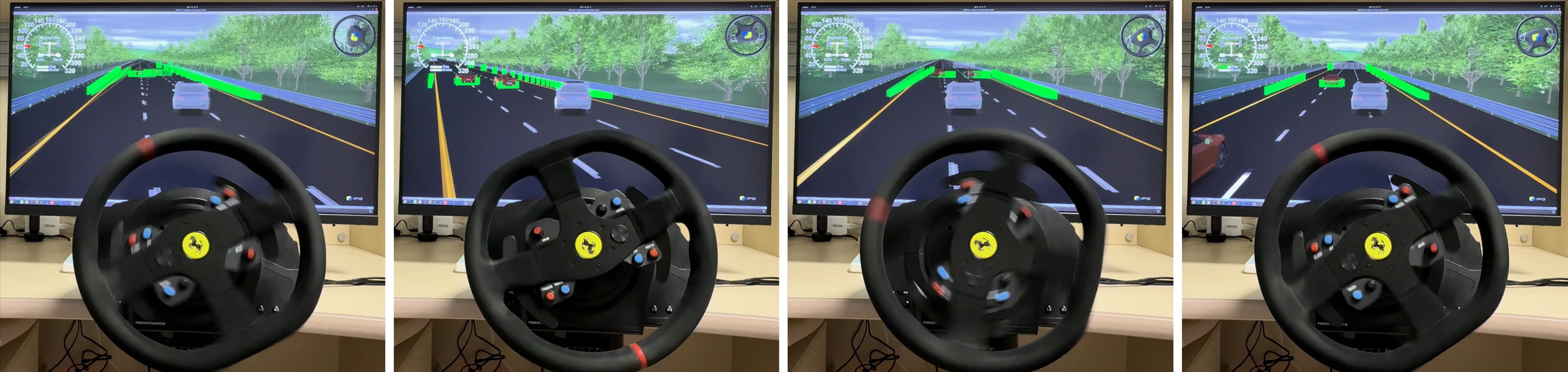}}
\caption{HILS validation setup for QC-SAC. (a) Overall HILS setup diagram. (b) Captured scene during the HILS validation. Videos of the HILS validation can be found at \url{https://youtu.be/w8qZLNeGQn8}.}
\label{fig:hils}
\end{figure}

The overall HILS architecture is illustrated in Fig.~\ref{fig:hils}. QC-SAC receives the vehicle and surrounding states from the simulator and outputs a steering rate command and a pedal command. The steering rate command is then transmitted to the steering wheel hardware, and the hardware controller physically drives the steering wheel to follow the commanded steering motion through a PID controller. After the steering wheel is actuated, the actual steering angle measured from the hardware is fed back to the simulator and applied to the CarMaker vehicle model. Therefore, the proposed policy is evaluated in a real-time, closed-loop manner that includes not only the empirical CarMaker vehicle and tire dynamics, but also the response of the physical steering hardware.

The HILS validation was conducted over 500 evaluation episodes. QC-SAC achieved a success rate of \(78.2\%\), with 391 successful episodes and 109 failed episodes. Compared with the software-only simulation result of \(81.8\%\) reported in TABLE~\ref{table:OCCA}, the success rate decreased by only 3.6 percentage points. This limited performance degradation indicates that the proposed policy can be executed through a physical steering hardware interface while maintaining performance close to the software-only simulation result. Although this result does not fully replace real vehicle experiments, it demonstrates that QC-SAC is not limited to a software-only simulation and can operate with real steering hardware feedback in a high-fidelity HILS environment.

\section{Conclusion} \label{sec:conclusion}

In this study, we introduced an innovative autonomous driving technology capable of safely controlling oversteer while avoiding surrounding obstacles, achieved through the proposed Q-Compared Soft Actor-Critic (QC-SAC). Developing an optimal driving policy for oversteer control and collision avoidance is particularly challenging with existing IL, RL, and HL techniques due to RL's reliance on random exploration and IL and HL's dependence on nearly optimal expert demonstrations, which are difficult to obtain. QC-SAC addresses these challenges by effectively utilizing immature demonstration data and rapidly adapting to novel situations, making it a robust solution for developing optimal driving policies. We validated the superior performance of the driving policy trained via QC-SAC in the proposed benchmark inspired by real-world driver training procedures, significantly outperforming conventional techniques. To provide realistic validation toward real-world implementation, the experiments were conducted using high-fidelity vehicle and tire models based on empirical data, and were further extended to HILS with physical steering hardware. Nevertheless, these evaluations cannot fully reproduce unpredictable real-world effects, particularly variations in tire-road friction, and a simulation-to-real gap therefore remains. Real-vehicle validation under diverse road conditions is an important direction for future work. Despite this remaining gap, the proposed framework demonstrates the feasibility of simultaneously controlling oversteer and avoiding collisions, providing an important step toward autonomous vehicle control in safety-critical driving situations.

\appendix

\subsection{Vehicle Model and Simulation Parameters}
\label{app:vehicle_model_parameters}

This appendix summarizes the vehicle model and simulation parameters used in IPG CarMaker, which were kept fixed across all experiments except for Appendix~\ref{sec:robustness}.

The simulated vehicle was based on the default BMW G30 pre-LCI 540i model provided in IPG CarMaker, with the main parameter settings summarized in Table~\ref{tab:app_vehicle_params}. The vehicle was configured as a rear-wheel-drive vehicle with an automatic torque-converter transmission. The engine, aerodynamic, and suspension characteristics were set using the default lookup table data provided in IPG CarMaker. The front and rear tire models were also configured using the empirical tire data provided in IPG CarMaker according to the corresponding tire sizes. Therefore, the experiments were conducted using empirical vehicle and tire data rather than a simplified equation-based vehicle model, reducing the discrepancy between the simulation environment and realistic vehicle behavior.

\begin{table*}[t]
\centering
\caption{Vehicle and simulation parameters used in IPG CarMaker.}
\label{tab:app_vehicle_params}
\begin{tabular}{llcl}
\toprule
Parameter & Symbol & Value & Unit \\
\midrule
Vehicle mass & \(m\) & 1600.0 & kg \\
Center of gravity position & \((x_{\mathrm{CG}},y_{\mathrm{CG}},z_{\mathrm{CG}})\)
& \((2.596,\;0.000,\;0.562)\) & m \\
Inertia tensor about CG & \(\mathbf{I}_{\mathrm{CG}} = \begin{bmatrix}
I_{xx} & I_{xy} & I_{xz} \\
I_{yx} & I_{yy} & I_{yz} \\
I_{zx} & I_{zy} & I_{zz}
\end{bmatrix}\) &
\(
\begin{bmatrix}
696.215 & 0.000 & -8.863 \\
0.000 & 2707.206 & 0.000 \\
-8.863 & 0.000 & 2986.624
\end{bmatrix}
\) & kg\,m\(^2\) \\
Gyration radius & \((r_x,r_y,r_z)\)
& \((0.660,\;1.301,\;1.366)\) & m \\
Front/rear axle load & \((W_f,W_r)\)
& \((799.731,\;800.269)\) & kg \\
Wheelbase & \(L\) & 2.975 & m \\
Distance from CG to front/rear axle & \((l_f,l_r)\)
& \((1.488,\;1.487)\) & m \\
Front tire size & -- & 245/35R19 & -- \\
Rear tire size & -- & 275/35R19 & -- \\
Rack travel to steering pinion angle & -- & 86.911 & rad/m \\
Road friction coefficient & \(\mu\) & 0.4 & -- \\
Kick plate yaw moment & \(T_{\mathrm{kp}}\) 
& \([-10000,\;-8000]\cup[8000,\;10000]\) & Nm \\
Kick plate duration & \(t_{\mathrm{kp}}\) 
& 0.3 & s \\
\bottomrule
\end{tabular}
\end{table*}

The oversteer scenario was generated by implementing a virtual kick plate, as described in Section~\ref{sec:benchmark}. At the beginning of each episode, the kick plate yaw moment \(T_{\mathrm{kp}}\) was randomly sampled from either \([-10000,-8000]\) or \([8000,10000]\)~Nm, thereby randomizing both the disturbance direction and magnitude. When the vehicle passed over the kick plate location, the virtual yaw moment \(T_{\mathrm{kp}}\) and the compensating lateral force \(F_{\mathrm{kp}}=-T_{\mathrm{kp}}/l_f\) were applied to the vehicle. This combination approximately cancels the moment about the rear axle center and induces a rear-end lateral kick, generating an oversteer motion similar to a real-world kick plate. The virtual yaw moment and lateral force were applied for \(0.3~\mathrm{s}\) and then released, while the road friction coefficient remained at \(\mu=0.4\) throughout the subsequent oversteer recovery and collision-avoidance maneuver.

\subsection{Sensitivity to Vehicle and Road Parameter Variations}
\label{sec:robustness}

The vehicle and simulation parameters summarized in Table~\ref{tab:app_vehicle_params} correspond to the nominal configuration used for training and standard evaluation. To further examine whether the proposed controller remains effective under deviations from this nominal configuration, we evaluate QC-SAC under variations of key vehicle and road parameters. We focus on three quantities that directly affect oversteer recovery: the vehicle mass \(m\), the yaw moment of inertia \(I_{zz}\), and the road friction coefficient \(\mu\). Each parameter is independently varied by \(\pm20\%\) from its nominal value, while all other vehicle, tire, and kick plate settings are kept fixed. For each parameter configuration, we conduct 100 evaluation episodes without retraining or fine-tuning QC-SAC and measure the control \& avoidance success rate. The deviation from nominal is computed with respect to the nominal success rate of 81.8\% reported in TABLE~\ref{table:OCCA}.

\begin{table}[t]
\centering
\caption{Sensitivity analysis of QC-SAC with respect to vehicle and road parameter variations.}
\label{table:robustness}
\begin{tabular}{lcc}
\toprule
Parameter & Success Rate & Deviation from Nominal \\
\midrule
$m$ ($-20\%$)      & 81\% & $-1.0\%$ \\
$m$ ($+20\%$)      & 74\% & $-9.5\%$ \\
$\mu$ ($-20\%$) & 37\% & $-54.8\%$ \\
$\mu$ ($+20\%$) & 82\% & $+0.2\%$ \\
$I_{zz}$ ($-20\%$) & 86\% & $+5.1\%$ \\
$I_{zz}$ ($+20\%$) & 69\% & $-15.6\%$ \\
\bottomrule
\end{tabular}
\end{table}

The results in TABLE~\ref{table:robustness} indicate that QC-SAC is most sensitive to large reductions in road friction. When $\mu$ decreases by $20\%$, the success rate drops to 37.0\% because the available lateral force becomes insufficient to recover from the induced oversteer even with maximum counter-steering. Increasing the yaw inertia also degrades performance, causing more frequent reverse-steer events due to yaw overshoot and delayed recovery. Except for these two adverse cases, QC-SAC remains effective under moderate variations around the training parameter setting, with less than 10\% performance degradation.

\subsection{Sensitivity to Reward Weights}
\label{sec:reward_sensitivity}

Following the robustness evaluation against vehicle and road parameter variations, we further examine the sensitivity of QC-SAC to the reward-weight settings in \eqref{eq:reward_total_OCCA}. The nominal reward weights are \(\lambda_1=0.8\), \(\lambda_2=0.2\), and \(\lambda_3=0.2\), corresponding to the safety, progress, and auxiliary reward terms, respectively. To isolate the effect of each reward component, each reward weight is varied independently while the other weights and all remaining training settings are kept fixed. Specifically, \(\lambda_1\) is varied from \(0.4\) to \(1.2\), while \(\lambda_2\) and \(\lambda_3\) are independently varied from \(0.1\) to \(0.3\). The vehicle model, tire model, and scenario configuration are kept identical to the nominal QC-SAC setting. Each trained policy is evaluated over 100 episodes under the nominal vehicle and road configuration, and the deviation from nominal is computed relative to the nominal success rate of 81.8\% reported in TABLE~\ref{table:OCCA}.

\begin{table}[t]
\centering
\caption{Sensitivity analysis of QC-SAC with respect to reward-weight variations.}
\label{table:reward_sensitivity}
\begin{tabular}{lccccc}
\toprule
Setting & \(\lambda_1\) & \(\lambda_2\) & \(\lambda_3\) & Success Rate & Deviation from Nominal \\
\midrule
\(\lambda_1\) low  & 0.4 & 0.2 & 0.2 & 78\% & $-4.6\%$ \\
\(\lambda_1\) high & 1.2 & 0.2 & 0.2 & 65\% & $-20.5\%$ \\
\(\lambda_2\) low  & 0.8 & 0.1 & 0.2 & 59\% & $-27.9\%$ \\
\(\lambda_2\) high & 0.8 & 0.3 & 0.2 & 73\% & $-10.8\%$ \\
\(\lambda_3\) low  & 0.8 & 0.2 & 0.1 & 38\% & $-53.5\%$ \\
\(\lambda_3\) high & 0.8 & 0.2 & 0.3 & 66\% & $-19.3\%$ \\
\bottomrule
\end{tabular}
\end{table}

TABLE~\ref{table:reward_sensitivity} shows that the performance of QC-SAC is affected by the relative balance among the reward components. Increasing the safety weight \(\lambda_1\) to 1.2 reduces the success rate, suggesting that an excessive safety penalty can make the policy overly conservative during transient oversteer recovery. Reducing the progress weight \(\lambda_2\) to 0.1 also decreases the success rate, indicating that sufficient progress incentive is required to complete the avoidance maneuver. The largest degradation occurs when the auxiliary weight \(\lambda_3\) is reduced to 0.1, showing that the stabilization-related auxiliary terms are also important for safe recovery. Increasing \(\lambda_3\) to 0.3 also degrades performance, suggesting that overly strong stabilization penalties can restrict corrective steering actions.

Overall, the results indicate that the nominal reward weights achieve the best balance among collision avoidance, forward progress, and vehicle stabilization in the considered scenario.

\subsection{Comparison with Physics-Based Trajectory-Tracking Controllers}
\label{sec:physics_based_comparison}

To position QC-SAC with respect to representative physics-based control approaches, we compare it with two physics-based trajectory-tracking yaw stability controllers constructed from a single-track bicycle model: a yaw stability controller based on a linear quadratic regulator (LQR-YSC) and a yaw stability controller based on model predictive control (MPC-YSC). Both controllers are designed on a linearized model of the lateral and yaw dynamics with the same tracking state
\[
x = [\, d_\mathrm{traj},\ \dot{d}_\mathrm{traj},\ \psi_\mathrm{traj},\ \dot{\psi},\ \beta \,]^\top,
\]
where \(d_\mathrm{traj}\) and \(\psi_\mathrm{traj}\) denote the lateral tracking error and heading error with respect to the reference trajectory, respectively, and the control input is the steering angle \(\delta\). To ensure a fair comparison with QC-SAC, the steering commands computed by both physics-based controllers are converted into steering rate commands before being applied to the CarMaker interface, with the same steering rate limit of \(700^\circ/\mathrm{s}\) used in QC-SAC. The pedal command is fixed to zero for both physics-based controllers. For LQR-YSC, the steering command at time step \(k\) is computed as \(\delta_k=-Kx_k\), and \(K\) is obtained by solving the Riccati equation for the linearized model. For MPC-YSC, the same trajectory-tracking problem is formulated as a finite-horizon optimal control problem, where a quadratic program is solved at each control step to minimize the predicted trajectory-tracking error over a prediction horizon of \(N=6\) steps, corresponding to \(0.3~\mathrm{s}\) with the simulation time step of \(0.05~\mathrm{s}\). These controllers provide representative model-based baselines under the same trajectory-tracking protocol, although their performance can depend on the model parameters, state representation, and tuning weights.

Designing a complete model-based trajectory planning module for the proposed benchmark would require solving a nonconvex trajectory optimization problem under highly nonlinear tire dynamics and complex obstacle configurations, and its performance would heavily depend on the particular optimization formulation and tuning choices. To avoid introducing such additional sources of variability, we instead exploit the fact that QC-SAC already produces dynamically feasible evasive maneuvers. For each randomly generated scenario in the oversteer control and collision avoidance benchmark, we first execute QC-SAC; whenever it successfully stabilizes the vehicle and avoids all obstacles, we record the resulting trajectory as a series of positions and headings. The same scenario configuration, including the obstacle layout and kick plate setup, is then reproduced for LQR-YSC and MPC-YSC, which are commanded to track the recorded trajectory as their reference. Internal states such as \(\beta\) and \(\dot{\psi}\) are not prescribed as references, so the tracking result reflects the intrinsic capability of the physics-based controllers rather than an idealized replay.

Across 100 trajectories recorded successfully by QC-SAC, LQR-YSC is able to complete the maneuver without spinning, leaving the drivable area, or colliding with obstacles in 71 episodes, whereas MPC-YSC succeeds in 47 episodes. This result indicates that the benchmark is challenging not only in terms of trajectory generation but also in terms of robust control under large side slip and nonlinear tire-force saturation. In particular, MPC-YSC achieves a lower success rate than LQR-YSC despite using finite-horizon prediction for the same trajectory-tracking problem. This result suggests that, in the large side slip regime induced by the kick plate, finite-horizon tracking based on a locally linearized model does not necessarily improve task-level robustness. The prediction model can become inaccurate under nonlinear tire force saturation and rapid yaw dynamics, and the short horizon optimization may produce delayed or overly conservative steering responses. These results empirically indicate that, in the oversteer control and collision avoidance benchmark, the proposed end-to-end learning-based framework, QC-SAC, achieves higher performance than the considered physics-based trajectory-tracking controllers.

\section*{Acknowledgments}
This work was supported by the National Research Foundation of Korea (NRF) grant funded by the Korea government (MSIT) (No. 2021R1A2C3008370). The authors thank IPG Automotive for providing CarMaker test licenses for the dissertation.

\bibliographystyle{IEEEtran}
\bibliography{reference}

\begin{IEEEbiography}
[{\includegraphics[width=1in,height=1.25in,clip,keepaspectratio]{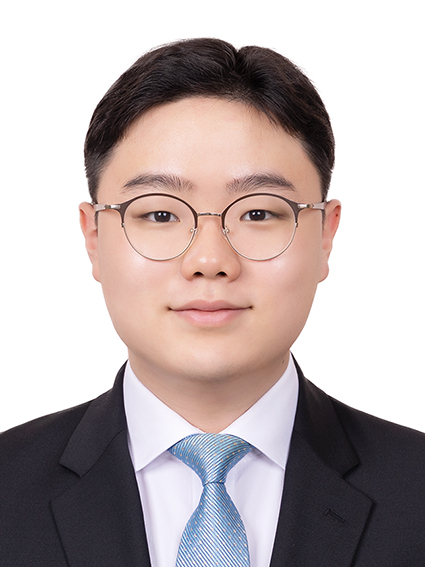}}]{Seokjun Lee} received a B.S. degree from the Department of Mechanical Engineering, Ulsan National Institute of Science and Technology (UNIST), Ulsan, South Korea, in 2023. He is currently enrolled in the Integrated M.S. and Ph.D. program at the Graduate School of Mobility, Korea Advanced Institute of Science and Technology (KAIST), Daejeon, South Korea. His research interests include end-to-end autonomous driving, autonomous driving on the handling limits, reinforcement learning and vehicle dynamics.
\end{IEEEbiography}

\begin{IEEEbiography}
[{\includegraphics[width=1in,height=1.25in,clip,keepaspectratio]{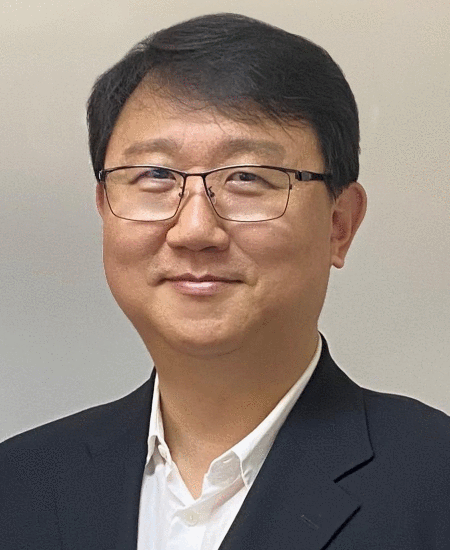}}]{Seung-Hyun Kong}(M’06–SM’16) is an Associate Professor in the CCS Graduate School of Mobility, Korea Advanced Institute of Science and Technology (KAIST), where he has been a faculty member since 2010. He received a B.S. degree in Electronics Engineering from Sogang University, Seoul, Korea, in 1992, an M.S. degree in Electrical and Computer Engineering from Polytechnic University (merged to NYU), New York, in 1994, and a Ph.D. degree in Aeronautics and Astronautics from Stanford University, Palo Alto, in 2005. From 1997 to 2004 and from 2006 to 2010, he was with companies, including Samsung Electronics (Telecommunication Research Center) and Nexpilot, both in Korea, Polaris Wireless in San Jose, and Qualcomm (Corporate R\&D) in San Diego, for advanced R\&D in mobile communication systems, wireless positioning, and assisted GNSS. His current research interests include deep neural networks for perception and localization using 4D Radar and sensor fusion, Reinforcement and Imitation learning algorithms for End-to-End autonomous driving, and neural network compression for embedded systems. He has authored more than 100 papers in peer-reviewed journals and conference proceedings and 14 patents, and his research group won the President’s Award in the 2018 International Student Autonomous Driving Competition from the Korean government. He served as a Program co-Chair of IEEE ITSC 2019, New Zealand, and the General Chair of IEEE IV 2024, Korea, and he has served as an Associate Editor of IEEE T-ITS since 2017.
\end{IEEEbiography}

\vfill

\end{document}